\documentclass[journal]{IEEEtran}
\ifCLASSINFOpdf
\else
\fi
\usepackage{cite}
\usepackage{amsmath,amssymb,amsfonts}
\usepackage{algorithmic}
\usepackage{graphicx}
\usepackage{multirow}
\usepackage{multicol}
\usepackage{textcomp}
\usepackage{hyperref}
\usepackage{filecontents}
\usepackage{booktabs}
\usepackage{lastpage}%
\usepackage{supertabular}%
\usepackage{tabu}%
\usepackage{diagbox}
\usepackage{longtable}%
\usepackage{color, colortbl}
\definecolor{Gray}{gray}{0.9}
\usepackage[algo2e,linesnumbered,ruled]{algorithm2e} 
\usepackage[numbers]{natbib}
\DeclareMathOperator*{\argmax}{argmax} 
\usepackage[flushleft]{threeparttable}
\begin{document}
\title{Genetic Programming-Based Evolutionary Deep Learning for Data-Efficient Image Classification}
	
\author{Ying Bi,~\IEEEmembership{Member,~IEEE}, Bing Xue,~\IEEEmembership{Senior Member,~IEEE}, and Mengjie Zhang,~\IEEEmembership{Fellow,~IEEE,}
\thanks{Manuscript received XXX; revised XXX;  revised XXX; accepted XXX. This work was supported in part by the Marsden Fund of New Zealand Government under Contracts VUW1913, VUW1914 and VUW2115, the Science for Technological Innovation Challenge (SfTI) fund under grant E3603/2903, MBIE Data Science SSIF Fund under the contract RTVU1914, and National Natural Science Foundation of China (NSFC) under Grant 61876169. (\emph{Corresponding author: Ying Bi})}

\thanks{The author Ying Bi is with School of Electrical and Information Engineering, Zhengzhou University, Zhengzhou 450001, China, and School of Engineering and Computer Science, Victoria University of Wellington, Wellington 6140, New Zealand (e-mail: ying.bi@ecs.vuw.ac.nz).} 

\thanks{The authors Bing Xue and Mengjie Zhang are with School of Engineering and Computer Science, Victoria University of Wellington, Wellington 6140, New Zealand (e-mail: bing.xue@ecs.vuw.ac.nz; mengjie.zhang@ecs.vuw.ac.nz).} 
\thanks{Color versions of one or more of the figures in this paper are available online at http://ieeexplore.ieee.org.}}
\markboth{IEEE transactions on xxxx,~Vol.~XX, No.~X, Month~year}
{Bi \MakeLowercase{\textit{et al.}}: paper title}
\maketitle
\begin{abstract}
Data-efficient image classification is a challenging task that  aims to solve image classification using small training data. Neural network-based deep learning methods are effective for image classification, but they typically require large-scale training data and have major limitations such as requiring expertise to design network architectures and having poor interpretability. Evolutionary deep learning is a recent hot topic that combines evolutionary computation with deep learning. However, most evolutionary deep learning methods focus on evolving architectures of neural networks, which still suffers from limitations such as poor interpretability. To address this, this paper proposes a new genetic programming-based evolutionary deep learning approach to data-efficient image classification. The new approach can automatically evolve variable-length models using many important operators from both image and classification domains. It can learn different types of image features from colour or gray-scale images, and construct effective and diverse ensembles for image classification. A flexible multi-layer representation enables the new approach to automatically construct shallow or deep models/trees for different tasks and perform effective transformations on the input data via multiple internal nodes. The new approach is applied to solve five image classification tasks with different training set sizes. The results show that it achieves better performance in most cases than deep learning methods for data-efficient image classification. A deep analysis shows that the new approach has good convergence and evolves models with high interpretability, different lengths/sizes/shapes, and good transferability.

\end{abstract}
\begin{IEEEkeywords}Evolutionary Deep Learning; Genetic Programming; Image Classification; Small Data; Evolutionary Computation; Deep Learning
\end{IEEEkeywords}
\IEEEpeerreviewmaketitle

\section{Introduction}\label{sec:introduction}
Image classification tasks have a wide range of applications such as detecting cancer from x-ray images, identifying a face from a set of photographs, and classifying fish species from underwater images \cite{schindelin2012fiji} \cite{schneider2012nih} \cite{lu2007survey} \cite{bi2021gpimage}. Although image classification has been investigated for decades, it remains a challenging task due to many factors, e.g., high inter-class similarity and intra-class dissimilarity, image distortion, and lack of sufficient training data. 
In recent years, data-efficient image classification that aims to use small data to perform effective image classification arise much attention. It can reduce the reliance on a large number of training data, therefore reducing the cost and effort for data collection, labelling, prepossessing, and storage. Data-efficient image classification is also important for many applications in medicine, remote sensing and biology, where the labelled data are not easy to obtain due to the cost, privacy and security.

Deep learning is a hot research area with many successful applications \cite{lecun2015deep}. The goal of deep learning is to automatically learn or discover multiple levels of abstraction from data (i.e. representation learning) that are effective for a particular task \cite{bengio2013representationlearning}. 
These methods often include three main characteristics, i.e., sufficient model complexity, layer-by-layer data processing, and feature transformation  \cite{zhou2018deep}. Deep learning methods include both neural-network(NN)-based methods such as convolutional NNs (CNNs) \cite{alzubaidi2021review} and non-NN-based methods such as deep forest \cite{zhou2018deep} and PCANet \cite{chan2015pcanet}. In recent years, deep CNNs become a dominant approach to image classification \cite{alzubaidi2021review}. However, these NN-based methods have major limitations, such as requiring expensive computing devices to run, a large number of training data to train, and rich expertise to design the architectures \cite{zhou2018deep}. 
Another important disadvantage of deep NNs is poor interpretability due to the large number of parameters and very deep structures in the model \cite{zhou2018deep}. 
To address these limitations, it is worth inventing new non-NN-based deep learning methods, which can not only maintain the diversity of the artificial intelligence/machine learning/computational intelligence research community but also bring new ideas to this community. 

Evolutionary deep learning (EDL) is a new research field that aims to use evolutionary computation (EC) techniques to evolve or optimise deep models \cite{zhan2022evolutionary} \cite{al2019survey}. 
The advantages, i.e., population-based beam search, non-differential objective functions, ease of cooperating with domain knowledge, robustness to dynamic changes, etc, have enabled EC methods to solve many complex optimisation and learning problems in various fields including finance, engineering, security, healthcare, manufacturing, and business \cite{al2019survey} \cite{back1997handbook}. 

Existing works of EDL can be broadly classified into two categories, i.e., the use of EC methods to automatically to optimise deep NNs by searching for architectures, weights, loss functions etc, and the use of EC methods to automatically evolve variable-length, relatively deep, non-NN-based models. In the first category, many EC methods such as genetic algorithms (GAs), genetic programming (GP), particle swarm optimisation (PSO), and evolutionary multi-objective optimisation (EMO) have been used to automatically evolve different types of deep NNs, e.g., CNNs, Autoencoders, recurrent neural networks (RNNs), and generative adversarial networks (GANs) \cite{zhan2022evolutionary} \cite{zhou2021survey} \cite{liu2020survey}. However, they often suffer from the limitations of deep NNs, such as requiring a large number of training examples, having “black-box” models with millions of parameters, and are computationally expensive. In the second category, to the best of our knowledge, the main method is GP, which can automatically evolve deep models to solve a problem \cite{al2019survey}. According to the definition of deep learning \cite{lecun2015deep} \cite{zhou2018deep}, GP is a deep learning method because it can automatically learn models with advanced operators that are sufficiently complex to perform data/feature transformation to solve tasks. There is also an increasing trend in recent works recognising GP as a deep learning method \cite{zhan2022evolutionary} \cite{martinez2021lights}. Unlike NN-based deep learning methods, GP-based EDL methods can automatically evolve variable-length models and their co-efficiencies without making any assumptions about the model structure/architecture \cite{langdon2013foundations}. Importantly, the models learned by GP often have varying complexity and potentially high interpretability, which are difficult to achieve in NNs. Benefiting from these advantages, GP-based EDL methods have been successfully applied to image classification with good performance \cite{bi2021gpimage} \cite{bi2021fewshot} \cite{bi2021multi}. However, the potential of GP has not been fully investigated. For tasks with small training data, limited computation resources and high requirements for model interpretability, where deep NNs have difficulty to address, it is necessary to investigate novel non-NN-based deep learning methods i.e. GP-based methods.

Existing GP-based EDL methods have limitations in solving image classification, which can be broadly summarised into three aspects. First, most existing methods have only been examined on datasets with gray-scale images. Real-world problems may have images with various numbers of channels such as RGB channels, so it is necessary to develop a new GP method applicable to these different images. Second, most of the existing GP-based methods have rarely been applied to large-scale image classification datasets such as SVHN, CIFAR10, CIFAR100, and ImageNet. Third, the potential of GP with different operators to automatically learn models has not been fully investigated. The flexible representation allows GP to automatically build models using different types of simple and complex operators from different domains as functions (internal nodes) \cite{bi2019tevc}. But there is still a lot of research potential in developing new representations including GP's tree structures, the function and terminal sets.

Therefore, this paper focuses on developing a new GP-based EDL approach to data-efficient image classification, where current NN methods can not provide satisfactory performance due to the requirement of large training data. To achieve this, we will develop a new GP-based EDL approach (i.e. EDLGP) with a new model/solution representation, a new function set and a new terminal set, enabling it to automatically learn features from gray-scale and/or colour images, select classification algorithms, and build effective ensembles for image classification. We will examine the performance of the new approach on well-known image classification datasets including CIFAR10, Fashion\_MNIST, and SVHN and two face image datasets under the scenario of small training data. EDLGP will be compared with a large number of existing competitive methods including CNNs of varying architectures and complexity. To highlight the potential of GP-based EDL methods, we perform a comprehensive comparison between EDLGP and CNNs for image classification. In addition, a deep analysis on the results in terms of convergence behaviour, tree/model size, model interpretability and transferability is conducted.

The main characteristics of the new EDLGP approach are summarised as follows.

\begin{enumerate}
   \item The EDLGP approach can evolve variable-length tree-based symbolic models, achieving promising classification performance in the data-efficient scenario, which are difficult for current popular NN-based methods to address due to the requirement of a large number of training images. The design of automatically evolving ensembles further enhance the generalisation, which is a key issue when the training set is small.
    
    \item Compared with existing methods, the EDLGP approach has a flexible multi-layer model representation to automatically evolve shallow or deep models for different image classification tasks. With this representation, EDLGP can deal with gray-scale and/or colour images, extract image features via different processes such as image filtering and complex feature extraction, perform classification by using cascading and ensembles with high diversity, and automatically select parameters for image operators and classification algorithms. Unlike NN-based methods, EDLGP evolves models with small numbers of parameters, which are easy to learn and less relying on the scale of the data.  
    
	\item The EDLGP approach can evolve models with high interpretability and transferability. This is important for providing insights into the problems being solved and also for future model reuse and development. 
	
\end{enumerate}

\section{Related Work}
This section reviews related works on evolutionary deep learning including automatically evolving CNNs and GP for image classification and related works on data-efficient image classification. The limitations of these works are summarised.

\subsection{Evolutionary Deep Learning (EDL)}
\subsubsection{NN-based EDL methods} 
The methods that fall into this category typically use EC techniques to optimise deep NNs by searching for architectures, weights, and other parameters. A detailed review of existing work can be found in \cite{zhan2022evolutionary} \cite{zhou2021survey} \cite{liu2020survey}. 
\citeauthor{sun2019evocnn} \cite{sun2019evocnn} proposed an evolutionary algorithm to search for CNN architectures and connection weights for image classification. In \cite{sun2019completely}, a GA-based method was proposed to automatically design block-based CNN architectures for image classification. \citeauthor{lu2020multiobjective} \cite{lu2020multiobjective} developed an EMO-based method (i.e. NSGA-II) to automatically search CNN architectures while optimising multiple objectives. 
\citeauthor{zhu2021real} \cite{zhu2021real} proposed an EC-based neural architecture search (NAS) method under the real-time federated learning framework. The computational costs are significantly reduced by using this method. \citeauthor{zhang2020efficient} \cite{zhang2020efficient} developed a method based on sampled training and node inheritance to improve the computational efficiency of EC-based NAS for image classification. 


\subsubsection{GP-based EDL methods} Unlike NNs, GP can use a flexible tree-like symbolic representation to automatically evolve trees/models with different lengths to meet the needs of problem solving. For many problems, including image classification, GP can automatically discover the representation of the data by evolving models of appropriate complexity. According to the definition of deep learning \cite{lecun2015deep}, GP can be a deep learning method because it can learn very complex models and transform the input data by using many internal nodes in the GP trees into higher-level data. 

An early review on GP-based EDL methods is presented in \cite{al2019survey}, which mentions typical GP methods such as 3-tier GP and multi-layer GP for image classification. These methods can automatically evolve models, extract features from raw images and perform classification, which is similar to CNNs. \citeauthor{rodriguez2018structurally} \cite{rodriguez2018structurally} developed a GP autoencoder with structured layers to achieve representation learning and introduced the concept of deep GP. \citeauthor{shao2014feature} \cite{shao2014feature} proposed a GP-based method that can use different filters and image operators to achieve feature learning for image classification. This method achieved better performance than CNNs on several datasets. \citeauthor{bi2021gpimage} presented a series of GP-based methods, such as FLGP \cite{bi2020effective}, FGP \cite{bi2019tevc}, and IEGP \cite{bi2020genetic}, for representation learning and image classification, where the evolved GP trees are constructed via multiple functional layers. 


\subsection{Data-Efficient Image Classification}
Data-efficient image classification focuses on solving image classification with a small number of available training data, which has become an increasingly important topic in recent years. \citeauthor{bruintjes2021vipriors} \cite{bruintjes2021vipriors} and \citeauthor{lengyel2022vipriors} \cite{lengyel2022vipriors} proposed the first and second well-known challenges on data-efficient deep learning, which aims to develop new deep learning methods using small training data to solve computer vision tasks including image classification, object detection, instance segmentation, and action recognition. In \cite{bruintjes2021vipriors} \cite{lengyel2022vipriors}, two important rules of solving data-efficient image classification are proposed, i.e., 1) models can only be trained from scratch using the training set of the task, 2) other data, transfer learning and model pre-training are prohibited.

\citeauthor{arora2019harnessing} \cite{arora2019harnessing} proposed a convolutional neural tangent kernel (CNTK) method, e.g., classifying CIFAR10 with 10 to 640 training images. The results showed that the CNTK methods with different numbers of layers can beat ResNet. \citeauthor{brigato2021close} \cite{brigato2021close} investigated different CNNs of varying complexity for image classification on small data. The results showed that dropout can improve the generalisation of CNNs. 
\citeauthor{bi2022using} \cite{bi2022using} proposed a multi-objective GP method to simultaneously optimise training accuracy and a distance measure to improve generalisation of the classification system using a small training set. 

\citeauthor{barz2020deep} \cite{barz2020deep} proposed the well-known Cosine loss for training deep NNs using small datasets. By maximising the cosine similarity between the outputs of the NNs and the one-hot vectors of the true class, this loss function gained better results than the commonly used cross-entropy loss on several image datasets. \citeauthor{brigato2021tune} \cite{brigato2021tune} proposed eight data-efficient image classification benchmarks including object classification, digit recognition, medical image classification, and satellite image classification. Several methods were investigated to solve these benchmarks. The results showed parameter tuning in terms of batch size, learning rate and weight decay can improve the performance of existing methods on these datasets. \citeauthor{sun2020visual} \cite{sun2020visual} proposed visual inductive priors framework with a new neural network architecture for data-efficient image classification. A loss function based on the positive class classification loss and the  intra-class compactness loss was developed to further improve the generalisation. This method ranked the first in the first data-efficient image classification challenge \cite{bruintjes2021vipriors}. \citeauthor{zhao2020distilling} \cite{zhao2020distilling} proposed a method with a two-stage manner to train a teacher network and a student network for data-efficient image classification. This method ranked the second in the challenge \cite{bruintjes2021vipriors}.


\subsection{Summary}
Existing work shows the potential of GP for data-efficient image classification \cite{bi2021fewshot} \cite{al2016binary} \cite{bi2020learning}.  However, all these methods focus on relatively simple tasks and use gray-scale images. Furthermore, very few GP-based methods have been tested on commonly used datasets, such as CIFAR10. The comparison between GP-based EDL methods and NN-based deep learning methods is not sufficient and comprehensive to show the advantages of GP in solving data-efficient image classification. To address these limitations and further explore the potential of GP-based EDL methods, this paper proposes a new data-efficient image classification approach based on GP and conducts comprehensive comparisons between the new approach and CNN to demonstrate its effectiveness. 

\begin{figure*}
	\centering
	\includegraphics[width=\linewidth]{ 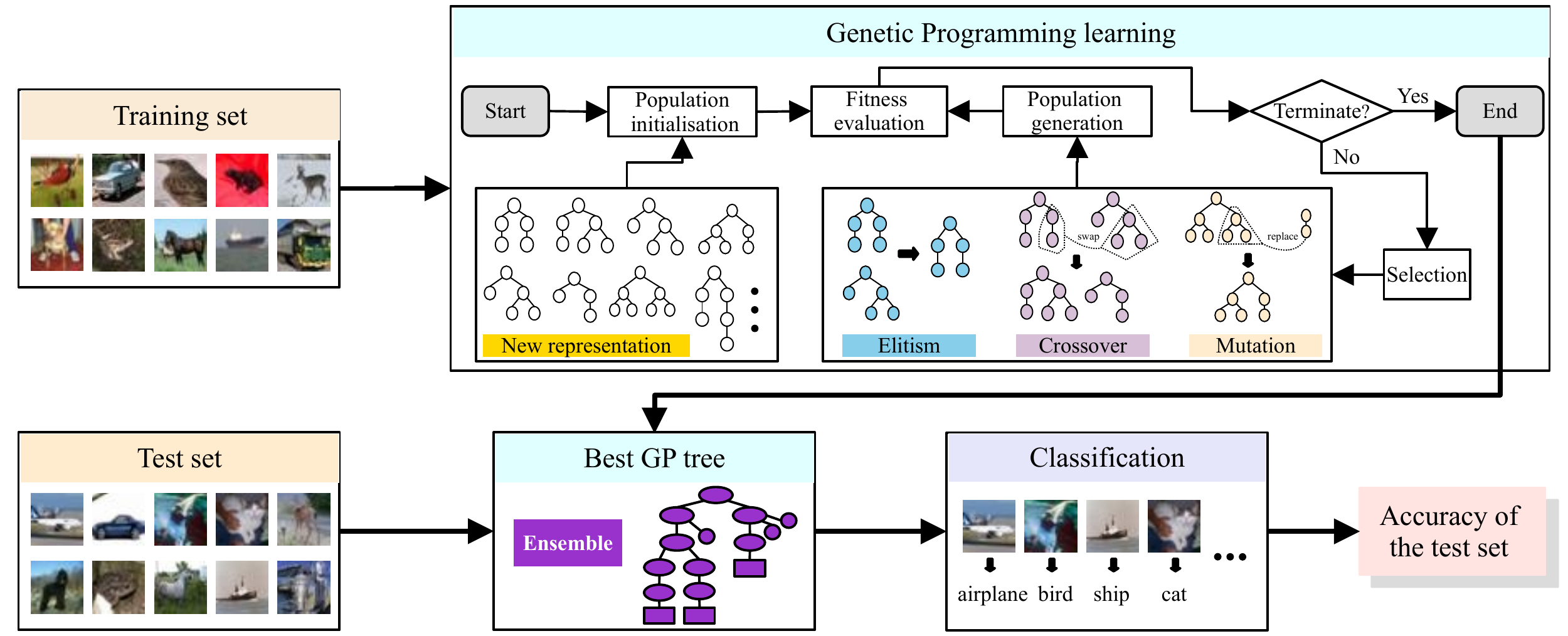}
			\vspace{-6mm}
	\caption{The overall algorithm of the proposed EDLGP approach to image classification.}
	\label{fig:overallalgorithm}
			\vspace{-4mm}
\end{figure*}

\section{The Proposed Approach}

This section introduce the proposed EDLGP approach to image classification, including the overall algorithm, the individual representation, genetic operators, and fitness function. To highlight the advantages of the new approach, a detailed comparison between it with deep CNNs is presented.

\subsection{Overall Algorithm}
An image classification task often consists of a training set and a test set. The training set is denoted as $\mathcal{D}_{train} = \{(\boldsymbol{x}_i, y_i)\}_{i=1}^m$ and the test set is denoted as $\mathcal{D}_{test} = \{\boldsymbol{x}_j\}_{j=1}^n$, where $\boldsymbol{x}\in \mathbb{R}^{G\times W\times H}$ denotes the images with a size of $W\times H$ and a channel of $G$, and $y \in \mathbb{Z}$ denotes the class label of the image (the total number of classes is $C$). The number of training images is $m$ and the number of testing images is $n$. The task is to use $\mathcal{D}_{train}$ to build a model $g(\boldsymbol{x})$ that can effectively predict $y$ for testing/unseen images $\mathcal{D}_{test}$.

The overall process of EDLGP for image classification is shown in Fig. \ref{fig:overallalgorithm}. The input of the system is the training set $\mathcal{D}_{train}$ and the output is the best GP tree/model $g(\boldsymbol{x})$.  EDLGP starts by randomly initialising a number of tree-based models according to the predefined representation, the function set and the terminal set. The population is evaluated using the fitness function, where each model/tree/individual is assigned with a fitness value i.e. the accuracy of the training set. During the evolutionary process, promising individuals are selected using a selection method and new individuals are generated using genetic operators i.e. subtree crossover and subtree mutation. A new population is created by copying a proportion of individuals with the highest fitness values and by generating new individuals from crossover and mutation operations. The process of population generation, fitness evaluation and selection is repeated until reaching a predefined termination criterion. After the evolutionary process, the best GP tree/model is returned. The best tree $g^*(\boldsymbol{x})$ is used to classify images in the test set.

The optimisation process of the EDLGP approach can be defined as follows
\begin{equation} \label{featurelearning}
g^*(\boldsymbol{x}) = \argmax_{f\in \mathcal{F}, ~t \in \mathcal{T}}~\mathcal{L}(g(\boldsymbol{x}),~\mathcal{D}_{train})
\end{equation}
where $g(\boldsymbol{x})$ represents a GP tree/model and $\mathcal{L}$ represents the objective/fitness function. $\mathcal{F}$ denotes the functions and $\mathcal{T}$ denotes the terminals, which are employed to construct GP models/trees. The best model is learned by using EDLGP with a goal of maximising $\mathcal{L}$ using the training set $\mathcal{D}_{train}$. Note that $\mathcal{L}$ can be a loss function as that in NNs to be minimised.

To find the best model $g^*(\boldsymbol{x})$, a new representation, a new function set, and a new terminal set are developed in EDLGP. These new components allow EDLGP to automatically evolve variable-length models that extract informative image features and build effective ensembles of classifiers for classification. 

\subsection{Components of EDLGP}
The EDLGP approach has a new model representation/encoding, making it different from existing GP methods \cite{bi2021gpimage} \cite{al2019survey}. Furthermore, EDLGP uses genetic operators to create new populations of trees and a fitness function for evaluating the new models. These components are introduced as follows.

\begin{figure*}
	\centering
	\includegraphics[width=\linewidth]{ 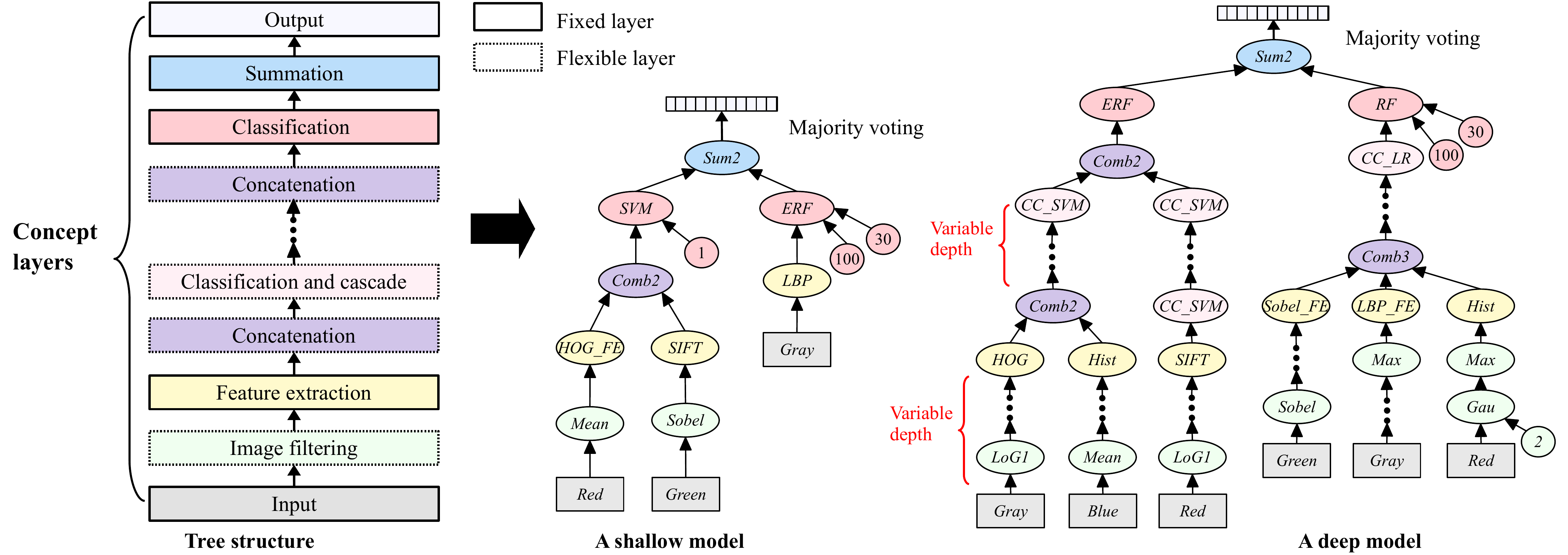}
		\vspace{-6mm}
	\caption{The tree structure of EDLGP and two example trees/models. The tree structure consists of a number of concept layers, where each layer uses a set of predefined functions that can be searched by EDLGP to build the model. The tree structure allows EDLGP to construct shallow and deep models (as the two examples) with different functions, sizes and shapes to solve different tasks. Each tree can extract features from raw images and construct ensembles of classifiers to achieve effective classification.}
	\label{fig:structureedlgp}
		\vspace{-4mm}
\end{figure*}

\subsubsection{New Model Representation}
EDLGP uses a tree-based structure based on strongly typed GP \cite{montana1995strongly} to represent the model for image classification. The tree structure and two example trees are shown in Fig. \ref{fig:structureedlgp}. Specifically, the tree structure consists of nine concept layers, i.e., input, image filtering, feature extraction, concatenation, classification and cascade, concatenation, classification, summation, and output. The input layer represents the inputs of a GP tree, such as raw images and the parameters of operators. The image filtering layer uses a set of commonly used image filters to process the input image. The feature extraction layer contains well-known image descriptors that can generate a set of features from images. The concatenation layers aim to concatenate the features generated from the previous layer to create a more comprehensive feature set. The classification and cascade layers perform classification and cascade the predicted class labels or weights to the input feature vector to form a stronger feature set. This idea was borrowed from cascade learning \cite{zhang2019deep}. The features generated by the classification and cascade layer can be further concatenated by a concatenation layer. The classification layer performs classification on the image using the features from its child nodes. The summation layer adds the predicted ``probabilities" from all its child nodes/classifiers for each class. The output layer performs the majority voting to make a prediction on which class the image belongs to, i.e. the class label. 

The tree structure of EDLGP consists of flexible layers and fixed layers with a good balance between the necessary model complexity of dealing with an image classification task and the flexibility of the model growing in depth. Specifically, the input, feature extraction, classification, summation, and output layers are fixed, which follows a commonly used manner for building an ensemble for image classification. These layers must appear in each GP tree/model, but the functions used in those layers vary with trees/models. The flexible layers are the image filtering layer, the concatenation layer, and the classification and cascade layer, which are optional for constructing GP trees/models. These flexible layers allow EDLGP to generate more effective features by using different numbers of functions as internal nodes in the GP trees. This is important for dealing with difficult image classification tasks that require sufficiently complex data transformation to generate informative features. 

There are \textbf{four main} characteristics of the new model representation. First, it allows EDLGP to evolve shallow and deep models for different tasks. A simple model may only have a few internal and leaf nodes, similar to the example tree shown in the middle part of Fig. \ref{fig:structureedlgp}. A deep model may be constructed by a large number of internal nodes, as the example tree is shown in the right part of Fig. \ref{fig:structureedlgp}. Second, this representation allows EDLGP to construct ensembles of classifiers/ensembles for image classification, which enhances the effectiveness of the models when the number of training images is small. Third, this representation allows the constructed model to automatically learn effective image features with different processes, i.e., image filtering, feature extraction based on commonly used image descriptors, and cascading after classification, which are common in computer vision and machine learning. Fourth, EDLGP can evolve ensembles with high diversity for image classification. The ensembles include cascade classifiers and classifiers built using features extracted by the corresponding subtree. The diversity ensures the effectiveness of the constructed ensembles. These characteristics make EDLGP significantly different from existing GP-based methods for image classification. 

\subsubsection{Functions and Terminals}
The functions represent the internal and root nodes and the terminals represent the leaf nodes of GP trees/models. Different from existing GP methods, EDLGP uses more functions to extract different types of features and new classification and cascade functions to evolve deep models according to the model representation.

All \textit{terminals} are listed in Table \ref{all_terminals}. The terminals include the input image to be classified and parameters for corresponding functions. If the input images are colour, the terminals will be Red, Blue, Green, and gray, representing three colour channels and the gray-scale channel. If the input images are gray-scale, the terminal will be gray. The remaining terminals are $t$, $d$, $f$, $\theta$, $o_1$, $o_2$, and $\sigma$, which have their value ranges according to commonly used settings. Their values are automatically selected during the evolutionary process. 

\begin{table}
		\vspace{-4mm}
	\footnotesize
	\setlength{\tabcolsep}{0.6em} 
	\caption{Terminals}
	\vspace{-4mm}
	\begin{center}
		\begin{tabular}{p{0.07\textwidth}p{0.38\textwidth}}
			\hline 
		\textbf{Terminal}&\textbf{Description}\\ \hline
    	\emph{Red}, \emph{Blue}, \emph{Green}  & The red, blue, and green channels of the colour image. Each of them is a 2D array with values in range [0, 1]\\
	    \emph{gray}  &The gray-scale image, which is a 2D array with the values in range [0, 1]\\
		$t$     &The number of trees in RF, ERF, CC\_RF, and CC\_ERF. It is an integer in range [50, 1000] with a step of 50\\
		 $d$     &The maximal tree depth in RF, ERF, CC\_RF, and CC\_ERF. It is an integer in range [10, 100] with a step of 10\\
		 $f$     &The frequency of Gabor and Gabor\_FE. Its value is in range $[\pi/8, \pi/2]$ with a step of $\pi /2{\sqrt{2}}$\\
		 $\theta$ &The orientation of Gabor. Its value is in range $[0, \frac{7\pi}{8}]$ with a step of $\frac{\pi}{8}$\\
		 $o_1, o_2$ &The derivative orders of GauD and GauD\_FE. It is an integer in range $[0, 2]$\\
		$\sigma$  &The standard deviation of Gau and Gau\_FE. It is an integer in range $[1, 3]$\\
			\hline
		\end{tabular}	
		\label{all_terminals}
	\end{center}
	\vspace{-6mm}
\end{table}

\begin{figure}[htbp]
	\vspace{-4mm}
	\centering
    \includegraphics[width=\linewidth]{ 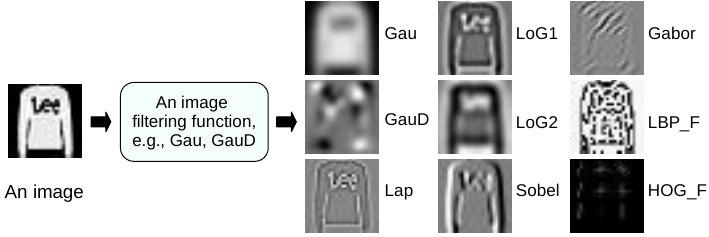}
	\vspace{-6mm}
	\caption{Illustration of image filtering and the images obtained by applying some functions.}
	\label{fig:image_filtering}
	\vspace{-4mm}
\end{figure}

The \textit{image filtering} layer includes commonly used image filtering operators, i.e., Mean, Median, Min, Max, Gau, GauD, Lap, LoG1, LoG2, Sobel, and Gabor. Functions at this layer also include the operators that can process an image, such as HOG\_F, LBP\_F, Sqrt, and ReLU. In addition, the functions include Add\_MaxP and Sub\_MaxP, which take two images as inputs, perform $2\times 2$ max-pooling to the images, add and subtract the two small images to generate a new image, respectively. If the size of the two input images is not the same, the two functions will perform max-pooling only on the small image to make the size the same before sum or subtraction. This is possible since the size of images can only be reduced by using the 2$\times$2 max-pooling operation in the function set. By using these different operators, it is expected to generate more effective features from images. Table \ref{image_filtering} lists all functions and the process is illustrated in Fig. \ref{fig:image_filtering}.

\begin{table}[htbp]
		\vspace{-4mm}
	\footnotesize
	\setlength{\tabcolsep}{0.6em} 
	\caption{Image filtering functions}
	\vspace{-4mm}
	\begin{center}
		\begin{tabular}{p{0.06\textwidth}p{0.39\textwidth}}
			\hline 
			\textbf{Function}&\textbf{Description}\\ \hline
						Mean		&Perform 3$\times$3 mean filtering \\	
			Median		&Perform 3$\times$3 median filtering\\
			Min	        &Perform 3$\times$3 minimum filtering\\
			Max	        &Perform 3$\times$3 maximum filtering \\
			Gau	 &Perform Gaussian filtering and the standard deviation is $\sigma$ \\		
			GauD		&Generate a new image by calculating derivatives of Gaussian filter with standard deviation $\sigma$ and orders $o_1$ and $o_2$\\		Lap		    &Perform Laplacian filtering\\
			LoG1	    &Perform Laplacian of Gaussian filtering, and the standard deviation is 1	\\
			LoG2		&Perform Laplacian of Gaussian filtering, and the standard deviation is 2\\
			Sobel	    &Perform 3$\times$3 Sobel filtering to the input image\\
		    Gabor	    &Perform Gabor filtering, where the orientation is $\theta$ and the frequency is $f$\\	
			LBP\_F		&Generate a LBP image\\
			HOG\_F		&Generate an HOG image\\			
			Sqrt		&Return sqrt root value of each value of the input image. Return 1 if the value is negative\\ 
            ReLU        &Perform the operation using rectified linear unit on the input image\\			
			Add\_MaxP   &Perform $2\times 2$ max-pooling on two input images or the smaller image and add the two images\\
			Sub\_MaxP   &Perform $2\times 2$ max-pooling on  two input images or the smaller image and subtract the two images\\
			\hline
		\end{tabular}	
		\label{image_filtering}
	\end{center}
\end{table}

\begin{figure}[htbp]
	\centering
    \includegraphics[width=\linewidth]{ 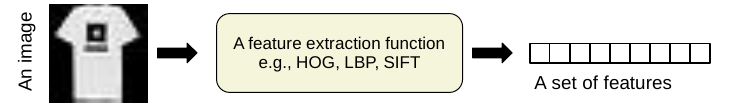}
     	\vspace{-6mm}
	\caption{Illustration of feature extraction.}
	\label{fig:feature_extraction}
	\vspace{-2mm}
\end{figure}

The \textit{feature extraction layer} contains 11 functions listed in Table \ref{feature_extraction} that can extract different types of features from images. The process is illustrated in Fig. \ref{fig:feature_extraction}. These functions include commonly used image descriptors, i.e., Histogram (Hist), HOG, LBP, HOG\_FE, LBP\_FE, and Gabor\_FE, and methods for detecting edges, i.e., Sobel\_FE and GauD\_FE, and others, i.e., Conca and Gau\_FE. The Hist, HOG and LBP functions extract features from the input image. The X\_FE functions perform corresponding filtering operations on the image and extract the features by concatenating the image into a vector. The Conca function can concatenate two images to form a feature vector.

\begin{table}[!h]
		\vspace{-4mm}
	\footnotesize
	\setlength{\tabcolsep}{0.6em} 
	\caption{Feature extraction functions}
	\vspace{-4mm}
	\begin{center}
		\begin{tabular}{p{0.07\textwidth}p{0.38\textwidth}}
			\hline 
			\textbf{Function}&\textbf{Description}\\ \hline
			Conca   &Concatenate two images into a feature vector\\
			Hist    &Extract 256 histogram features\\
			HOG     &Extract HOG features from the input image \cite{dalal2005histograms}. The features are the mean values of all $4\times 4$ grids from the image\\
			LBP         &Extract uniform LBP features \cite{ojala2002multiresolution}. The radius is 1.5 and the number of neighbours is 8 in LBP.\\
		    SIFT		&Extract 128 dense SIFT features from the input image \cite{vedaldi2010vlfeat}\\
		    LBP\_FE, HOG\_FE      &Generate a LBP or HOG image and concatenate the image into a feature vector\\
            Sobel\_FE, Gabor\_FE, Gau\_FE, GauD\_FE&Use the Sobel, Gabor, Gau, or GauD filter to process the input image and concatenate the new image into a feature vector\\
			\hline
		\end{tabular}	
		\label{feature_extraction}
	\end{center}
	\vspace{-4mm}
\end{table}

\begin{figure}[!htbp]
	\vspace{-2mm}
	\centering
	\includegraphics[width=\linewidth]{ 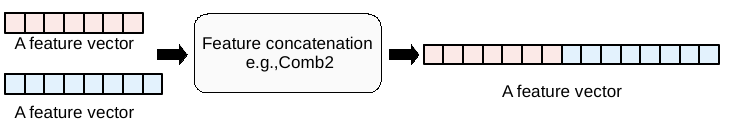}
		\vspace{-6mm}
	\caption{Illustration of feature concatenation using function Comb2.}
		\label{fig:feature_concatenation}
	\vspace{-4mm}
\end{figure}

The \textit{feature concatenation} layer uses Comb2, Comb3 and Comb4 to further concatenate the feature vectors from their 2, 3, and 4 child nodes into a feature vector, respectively, which increases the comprehension of the features for describing an image. This layer can increase the width of the GP trees and produce more features for the classification layer. The process is illustrated in Fig. \ref{fig:feature_concatenation}.

\begin{figure}[htbp]
	\vspace{-6mm}
	\centering
	\includegraphics[width=\linewidth]{ 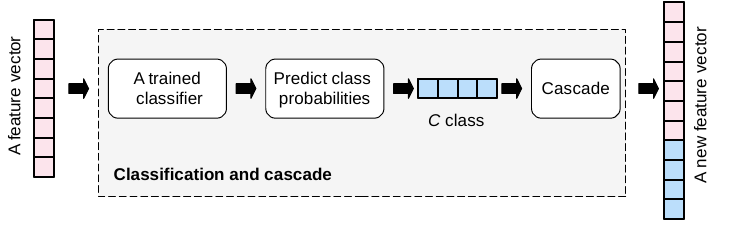}
	\vspace{-8mm}
	\caption{Illustration of classification and cascade.}
	\label{fig:classification_cascade}
	\vspace{-2mm}
\end{figure}

The \textit{classification and cascade} layer has four functions listed in Table \ref{reamining_functions}, i.e., cascade random forest (CC\_RF), cascade extremely randomised trees (CC\_ERF), cascade logistic regression (CC\_LR), and cascade support vector machine (CC\_SVM), which perform classification and concatenate the predicted class probabilities with the input features to form the output features, as shown in  Fig. \ref{fig:classification_cascade}. This follows the concept of cascade ensemble learning \cite{zhou2018deep}. The prediction of an instance is a $C$-dimensional vector denoting the probabilities for a $C$-class problem. If the number of features is $f_n$, the number of output features from these functions will be $f_n+C$. Unlike CC\_RF, CC\_ERF and CC\_LR, which are soft classifiers, CC\_SVM is a hard classifier so that the prediction vector is binary (discrete). To build effective classifiers, the main parameters of CC\_RF and CC\_ERF, i.e., the number of trees and the maximal tree depth, are set as terminals and their values can be automatically selected by EDLGP.

The \textit{classification} layer contains functions, i.e., RF, ERF, LR, and SVM, which take features as inputs and return the predicted class probabilities. The \textit{summation} layer has three functions, i.e., Sum2, Sum3 and Sum4, which take two, three and four vectors of predicted probabilities as inputs and add them, respectively. Each vector denotes the probabilities of all classes an instance belongs to. These functions are listed in Table \ref{reamining_functions}. The \textit{output} layer can make predictions according to the output of GP trees by assigning the class label with the highest probability to the input image.

\begin{table}[!h]
		\vspace{-4mm}
	\footnotesize
	\setlength{\tabcolsep}{0.6em} 
	\caption{Classification and cascade, classification and summation functions}
	\vspace{-4mm}
	\begin{center}
		\begin{tabular}{p{0.06\textwidth}p{0.4\textwidth}}
			\hline 
			\textbf{Function}&\textbf{Description}\\ \hline
			CC\_RF, CC\_ERF, CC\_LR, CC\_SVM	&Use the RF, ERF, LR, and SVM methods to perform classification. The outputs of these functions are the concatenation of the input features and the predicted class probabilities\\
			
			\hline
			RF      &Random forest classification method. The number of trees is $t$ and the maximal tree depth is $d$\\ 
		    ERF		&Extremely randomised trees classification method. The number of trees is $t$ and the maximal tree depth is $d$\\ 	
		    LR		&Logistic regression classification method \\ 	
		    SVM		&Support vector machine classification method\\	
			\hline
			Sum2/3/4	&Sum 2/3/4 vectors of predicted probabilities\\ 
			\hline	
		\end{tabular}	
		\label{reamining_functions}
	\end{center}
	\vspace{-4mm}
\end{table}

Besides the above functions/operators, it is possible to use other different functions at the corresponding layer in EDLGP. In other words, EDLGP can be easily extended by using more functions. The representation and the search mechanism allow EDLGP to automatically select the functions and terminals to build trees/models. However, it is suggested to carefully set the function set to make a good balance between the effectiveness of the potential models and the search space.


\subsubsection{Genetic Operators}
\begin{figure}[htbp]
	\centering
	\vspace{-4mm}
	\includegraphics[width=0.9\linewidth]{ 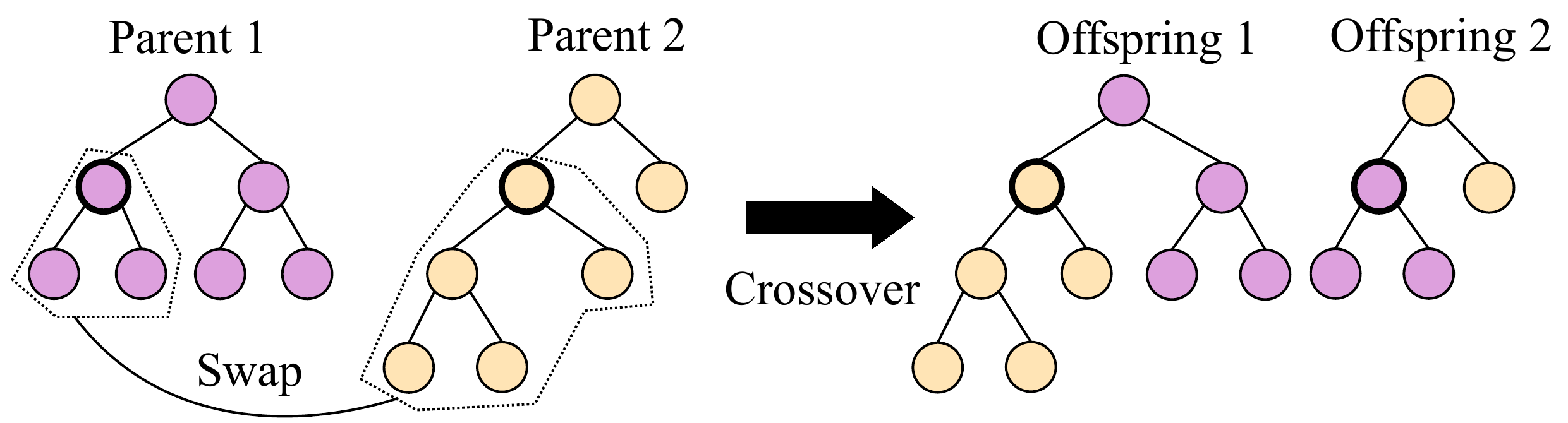}
	\vspace{-4mm}
	\caption{Illustration of crossover operation.}
	\label{fig:crossover}
		\vspace{-4mm}
\end{figure}

\begin{figure}[htbp]
	\centering
	\includegraphics[width=0.9\linewidth]{ 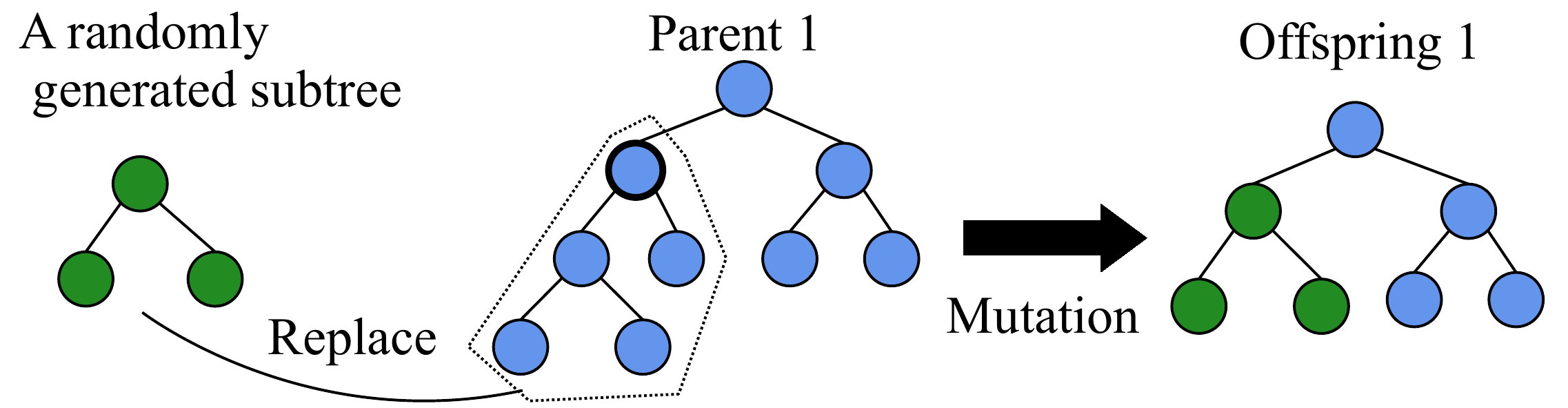}
	\vspace{-4mm}
	\caption{Illustration of mutation operation.}
	\label{fig:mutation}
		\vspace{-2mm}
\end{figure}

During the evolutionary process, new GP trees are generated using genetic operators, i.e., subtree crossover and subtree mutation. The subtree crossover operator is conducted on two selected trees/parents. It randomly selects two subtrees from the parents and swaps the two subtrees to generate two new trees, as shown in Fig. \ref{fig:crossover}. Note that the two selected subtrees must have the same output types in order to generate two feasible trees. The subtree mutation operator is conducted on one selected tree/parent. It randomly selects a subtree of the parent and replaces the subtree with a randomly generated subtree, as shown in Fig. \ref{fig:mutation}. 

\subsubsection{Fitness Evaluation} A fitness function is used in the fitness evaluation process to evaluate the goodness of the GP trees/models to the tasks. For a classification task, the most commonly used fitness function is the classification accuracy to be maximised, which is defined as 
\begin{equation} \label{fitness_function} 
\mathcal{L} = \frac{N_{correct}}{N_{total}}\times 100\%
\end{equation}
where $N_{correct}$ denotes the number of correctly classified instances and $N_{total}$ denotes the total number of instances in the training set. The fitness function is calculated using k-fold cross-validation on the training set. In EDLGP, the value of $K$ is set to $min(3, nc)$, where $nc$ denotes the number of instances in the smallest class of the training set. If $nc=1$, the fitness function will be the training accuracy. Note that this fitness function is effective for balanced classification. For unbalanced classification, other measures such as balanced accuracy can be used as the fitness function for EDLGP.

\subsection{Comparisons Between EDLGP and CNN}
The proposed EDLGP approach can automatically learn features and evolve ensembles for image classification. To highlight the characteristics of EDLGP, this section provides detailed analysis and comparisons between EDLGP (as a typical example of GP-based EDL methods) and CNNs, which is a dominant deep learning method for image classification. 

The main \textbf{similarities} are summarised as follows.

\begin{enumerate}
\item CNNs and EDLGP can automatically learn features and perform classification from raw images. 
\item CNNs and EDLGP are learning algorithms that learn models of many layers, which can perform complex data transformation to achieve good performance.
\item CNNs and EDLGP use image operations such as convolution and pooling to learn features from raw images.
\end{enumerate}

The main \textbf{differences} are summarised as follows.

\begin{enumerate}
    \item EDLGP uses a variable-length tree-based structure to represent a model, while CNNs typically uses a fixed-length layer-wise structure to represent a model. The flexible representation allows EDLGP to automatically learn models with different shapes, sizes, and depths. In contrast, CNNs need to predefine the model structure and length before training. 
    \item EDLGP is open to incorporate domain knowledge in a way of adding functions/operators to the corresponding layers, which cannot be easily achieved in CNNs.
    \item EDLGP has fewer model parameters and running parameters than CNNs. Detailed comparisons in terms of the running parameters are listed in Table \ref{parameters_EDLGP_CNN}. The number of CNN model parameters depends on the architecture configuration, which is often much larger than that of the automatically evolved EDLGP models.
    \item The models evolved by EDLGP have potentially higher interpretability than CNN models. The EDLGP model is composed of functions/operators from the image and machine learning domains, which can provide domain knowledge. The CNN model is composed of layers of a huge number of parameters, which are not straightforwardly to explain.
    \item The feature extraction and classification processes of CNNs and EDLGP are different. CNNs usually extract and construct features via a number of convolutional and pooling layers with learned kernels, and perform classification using softmax. EDLGP extracts features using image operators with predefined kernels and uses a traditional classification algorithm to perform classification. This makes CNNs and EDLGP applicable and effective for image classification under different scenarios. CNNs requires sufficient training data to learn effective feature maps and sub-sampling maps in these predefined layers to achieve promising performance, i.e., suitable for large-scale image classification. The EDLGP approach is more effective than CNNs when the training data is small, i.e., data-efficient image classification.
    \item The CNN methods can greatly benefit from running on GPU, while EDLGP would not at the current stage.
\end{enumerate}

\begin{table}
	\footnotesize
	\setlength{\tabcolsep}{0.3em} 
	\caption{Comparisons of running parameters of EDLGP and CNNs}
	\vspace{-6mm}
	\begin{center}
	\begin{tabular}{p{0.22\textwidth}p{0.25\textwidth}}
	\hline 
	\textbf{Convolutional neural networks}&\textbf{Evolutionary deep learning based on genetic programming}\\ \hline
	\textbf{Type of activation functions}: &\textbf{Type of selection}:\\
	~~ReLU, tanh, sigmoid, etc&~~Tournament selection or Roulette wheel selection\\
	\textbf{Architecture configurations}:& \textbf{Architecture configurations}:\\
	    ~~No. hidden layers&~~Function set\\
	    ~~No. convolutional layers&~~Terminal set\\
	    ~~No. pooling layers& \textbf{Optimisation configurations}:\\
	    ~~No. other layers&~~No. generations\\
	    ~~How to connect these layers&~~Population size\\
	   ~~No. features maps/nodes&~~Probability of elitism/reproduction\\
	   ~~Kernel sizes&~~Probability of mutation\\
	\textbf{Optimisation configurations}:&~~Probability of crossover\\
    ~~Learning rate&~~minimal and maximal tree size\\
    ~~Dropout: {0.25/0.50} &~~selection size\\
    ~~Momentum&\\
    ~L1/L2 weight regularisation penalty&~Tree generation method: full, grow, ramped-half-and-half \\
    ~Weight initialisation: uniform, Xavier Glorot, etc.& \\ 
    ~~Batch size: {32/64/128}& \\
	\hline
	\end{tabular}	
	\label{parameters_EDLGP_CNN}
	\end{center}
	\vspace{-4mm}
\end{table}

\section{Experiment Design}
This section describes the experiment design, including datasets, the comparison methods, and parameter settings. 

\subsection{Image Classification Datasets}
\begin{figure}[htbp]
	\centering
	\vspace{-2mm}
	\includegraphics[width=\linewidth]{ 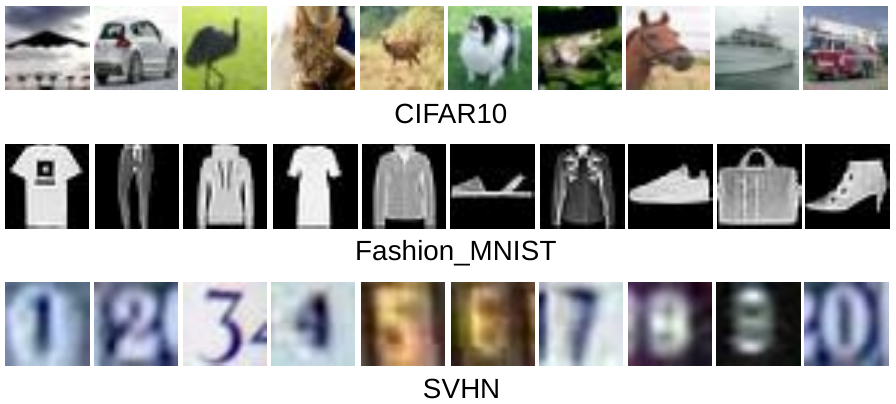}
	\vspace{-6mm}
	\caption{Example images of the CIFAR10, Fashion\_MNIST, and SVHN datasets. Each image represents one class.}
	\label{fig:datasets_popular}
		\vspace{-2mm}
\end{figure}

\begin{figure}
	\centering
	\includegraphics[width=\linewidth]{ 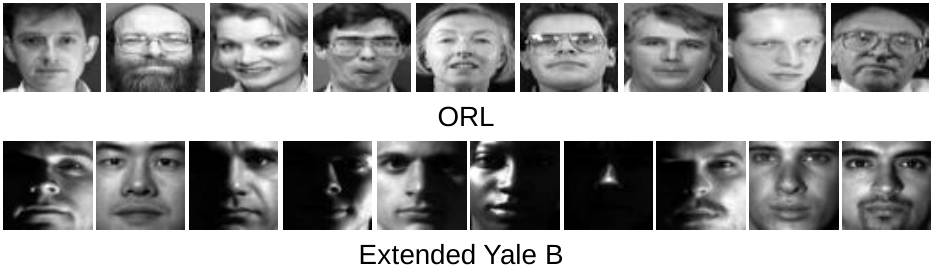}
	\vspace{-6mm}
	\caption{Example images of the ORL and Extended Yale B datasets.}
	\label{fig:datasets_att}
	\vspace{-2mm}
\end{figure}

To evaluate the effectiveness of the new approach, five different datasets are used, which are CIFAR10 \cite{krizhevsky2009learning}, Fashion\_MNIST (FMNIST in short) \cite{xiao2017fashion}, SVHN \cite{netzer2011reading}, ORL \cite{samaria1994parameterisation}, and Extended Yale B \cite{lee2005acquiring}. The CIFAR10, FMNIST, and SVHN datasets are object classification tasks. The ORL and Extended Yale B datasets are face classification tasks. The example images of these datasets are shown in Figs. \ref{fig:datasets_popular} and \ref{fig:datasets_att}. 

The CIFAR10 dataset is composed of 50,000 training images and 10,000 testing images. The images are colour and of size 32$\times$32. The Fashion\_MNIST dataset has 60,000 training images and 10,000 testing images in gray-scale. The image size is 28$\times$28. The SVHN dataset contains 73,257 colour images for training and 26,032 colour images for testing. The image size is 32$\times$32. The ORL dataset consists of 400 facial images from 40 people, i.e. 10 images per person. The Extended Yale B dataset has 2,424 facial images in 38 classes. 

\begin{table}[htbp]
	\vspace{-2mm}
	\caption{Summary of the datasets}
	\vspace{-4mm}
	\begin{center}
		\begin{tabular}{|p{0.215 \linewidth}|p{0.08\linewidth}|p{0.13\linewidth}|p{0.17\linewidth}|p{0.16\linewidth}|}
			\hline 
Dataset&\#Class &Image size&Train set size&Test set size\\
\hline	
CIFAR10&10&32$\times$32$\times$3&50,000&10,000\\ \hline
Fashion\_MNIST&10&28$\times$28&60,000&10,000\\ \hline
SVHN&10&32$\times$32$\times$3&73,257&26,032\\ \hline
ORL&40&56$\times$46&\multicolumn{2}{l|}{400}\\ \hline
Extended Yale B&38&32$\times$32&\multicolumn{2}{l|}{2424}\\ 
			\hline
		\end{tabular}	
		\label{table:datasets_new}
	\end{center}
	\vspace{-4mm}
\end{table}

%
%

The EDLGP approach is examined on data-efficient image classification problems. A small number of training images are used in the experiments. For the CIFAR10, Fashion\_MNIST and SVHN datasets, 10, 20, 40, 80, 160, 320, 640, and 1280 images are randomly selected from each class in the original training set to form a small training set used in the experiments, following the settings in \cite{brigato2021close}. The small datasets are called sCIFAR10-X, sFMNIST-X and sSVHN-X, where X denotes the number of training images in each class. To comprehensively show the performance, we also test EDLGP using very small training sets of CIFAR10, i.e., 1, 2, 4, 8, 16, 32, 64, and 128 images per class, following the settings in \cite{arora2019harnessing}. In all these settings, \emph{the original test sets are used for testing}. For the ORL dataset, 2, 3, 4, and 5 images are randomly selected to form the training sets and the remaining images are used for testing in the experiments \cite{liao2021deep}, respectively. For the Extended Yale B dataset, 15, 20, 25, and 30 images are randomly selected for training and the rest are used for testing \cite{zhang2019deep}, respectively. As a result, a large number of experiments are conducted to comprehensively demonstrate the performance of EDLGP in comparison with different deep learning methods. 

\subsection{Comparison Methods}
We compare EDLGP with existing methods including CNNs with different architectures, common methods, and non-NN-based deep learning methods, which are very effective methods on the corresponding datasets.

\begin{itemize}
    \item On sCIFAR10, sFMNIST and sSVHN, the comparisons methods are CNN-lc \cite{brigato2021close}, CNN-mc \cite{brigato2021close}, CNN-hc \cite{brigato2021close}, and ResNet-20 \cite{he2016deep}. CNN-lc, CNN-mc and CNN-hc denote different CNN methods with low, middle and high complexity used for image classification. 
    Different Dropout rates are investigated in these methods and the details can be found in \cite{brigato2021close}. 
    
    \item On few-shot sCIFAR10, the comparison methods are ResNet \cite{he2016deep}, 5-layer convolutional neural tangent kernel (CNTK) networks \cite{arora2019harnessing}, 8-layer CNTK \cite{arora2019harnessing}, 11-layer CNTK \cite{arora2019harnessing}, 14-layer CNTK \cite{arora2019harnessing}, and Harmonic Networks \cite{ulicny2019harmonic222}. 
    
    \item On ORL, the comparisons methods are Eigenface based on PCA \cite{turk1991eigenfaces}, Fisherface based on LDA \cite{belhumeur1997eigenfaces}, Laplacianface \cite{he2005face}, neighbourhood preserving embedding (NPE) \cite{he2005neighborhood}, marginal Fisher analysis (MFA) \cite{yan2006graph}, CNN \cite{liao2021deep}, and deep metric learning based on CNN and k nearest neighbour classification (KCNN) \cite{liao2021deep}.
    
    \item On Extended Yale B, the comparisons methods are collaborative representation classifier (CRC) \cite{zhang2011sparse}, SRC \cite{wright2008robust}, correntropy-based sparse representation (CESR) \cite{he2010maximum}, robust sparse coding (RSC) \cite{yang2011robust}, half-quadratic with the additive form (HQA) \cite{he2013half}, half-quadratic with the multiplicative form (HQM) \cite{he2013half}, NMR \cite{yang2016nuclear}, robust matrix regression (RMR) \cite{xie2017robust}, Fisher discrimination dictionary learning (FDDL) \cite{yang2014sparse}, low-rank shared dictionary learning (LRSDL) \cite{vu2017fast}, DCM based on NMR (DCM(N)) \cite{zhang2019deep}, and DCM based on SRC (DCM(S)) \cite{zhang2019deep}. 
\end{itemize}

The parameter settings of these comparison methods refer to the corresponding references. The other experimental settings refer to  \cite{zhang2019deep} \cite{brigato2021close} \cite{liao2021deep}. The goal of this paper is to explore the potential of GP-based EDL for data-efficient image classification. The aforementioned methods are used for comparisons because they are effective methods on these datasets using small training sets. We also compare the EDLGP approach with more state-of-the-art data-efficient deep learning methods mentioned in \cite{brigato2021tune} on CIFAR10 under different settings. Due to the page limit, the results and analysis is presented in the supplementary materials. We do not include GP-based methods for comparisons because they have not been applied or not designed for solving these datasets.

\subsection{Parameter Settings}
 
The parameter settings for the EDLGP approach follow the commonly used settings in the GP community \cite{bi2021gpimage} \cite{bi2019evolutionary}. The maximal number of generations is set to 50 and the population size is set to 100. The crossover rate is 0.5, the mutation rate is 0.49, and the elitism rate is 0.01. The selection method is tournament selection with size 5. The population initialisation method is the ramped-half-and-half method. The initial tree depth is 2-10. The maximal tree depth is 10. Note that these parameter values for EDLGP can be fine-tuned on each dataset for optimal settings. However, our study aims to investigate a more general approach that can achieve good performance using a common setting. In the experiments, EDLGP is executed 10 times independently on each dataset using different random seeds because of the stochastic nature and the high computation cost.

\section{Results and Discussions}
This section compares the performance of the EDLGP approach with the benchmark methods on different datasets of various numbers of training images. It also deeply analyses different aspects of EDLGP in terms of convergence behaviours, interpretability and transferability of models/trees. 



\begin{table*}[!htbp]
\footnotesize
	\setlength{\tabcolsep}{0.4em} 
	\caption{Mean Classification accuracy (\%) of CNNs of varying complexity and classification accuracy (\%) of EDLGP on small CIFAR10, Fashion\_MNIST and SVHN datasets of varying training set size}
	\vspace{-6mm}
	\begin{center}
		\begin{tabular}{llllllllll}
			\hline 
		Model&Dropout&sCIFAR10-10&sCIFAR10-20&sCIFAR10-40&sCIFAR10-80&sCIFAR10-160&sCIFAR10-320&sCIFAR10-640&sCIFAR10-1280\\
		CNN-lc&0.0&27.1&32.4&36.1&41.8&45.3&50.1&54.1&59.4\\
		CNN-lc&0.4&29.7&33.8&38.2&43.5&47.1&52.8&57.2&61.7\\
		CNN-lc&0.7&29.7&34.9&40.0&44.9&49.4&53.9&58.1&62.3\\
		CNN-mc&0.0&28.5&34.3&38.8&43.0&48.6&53.7&58.8&63.3\\
		CNN-mc&0.4&29.7&34.9&39.9&45.4&50.9&55.8&61.0&66.2\\
		CNN-mc&0.7&31.5&36.2&41.3&47.1&51.9&57.7&62.8&67.6\\
		CNN-hc&0.0&30.1&34.2&39.1&44.7&50.6&56.1&61.2&65.9\\
		CNN-hc&0.4&31.7&36.1&40.8&46.5&52.0&58.0&63.5&68.8\\
		CNN-hc&0.7&31.9&37.0&42.5&48.1&53.9&59.5&\textbf{64.8}&69.8\\
		ResNet-20&--&23.3&29.0&31.9&38.5&44.7&51.3&62.3&\textbf{71.5}\\
	\cellcolor[gray]{0.9}EDLGP(best)&\cellcolor[gray]{0.9}--&	\cellcolor[gray]{0.9}\textbf{35.8}&\cellcolor[gray]{0.9}\textbf{43.2}&	\cellcolor[gray]{0.9}\textbf{48.1}&\cellcolor[gray]{0.9}\textbf{52.3}&	\cellcolor[gray]{0.9}\textbf{58.2}&\cellcolor[gray]{0.9}\textbf{60.6}&\cellcolor[gray]{0.9}\textbf{64.8}&\cellcolor[gray]{0.9}64.5\\
	\cellcolor[gray]{0.9}EDLGP(mean)&\cellcolor[gray]{0.9}--&\cellcolor[gray]{0.9}\textbf{31.7$\pm$3.8}
	&\cellcolor[gray]{0.9}\textbf{37.7$\pm$4.0}&\cellcolor[gray]{0.9}\textbf{45.7$\pm$1.6}&\cellcolor[gray]{0.9}\textbf{49.3$\pm$1.8}&\cellcolor[gray]{0.9}\textbf{54.3$\pm$2.5}&\cellcolor[gray]{0.9}58.7$\pm$1.5&\cellcolor[gray]{0.9}61.6$\pm$2.0&\cellcolor[gray]{0.9}61.1$\pm$3.4\\
\hline
Model&Dropout&sFMNIST-10&sFMNIST-20&sFMNIST-40&sFMNIST-80&sFMNIST-160&sFMNIST-320&sFMNIST-640&sFMNIST-1280\\
		CNN-lc&0.0&71.1&74.0&77.8&81.0&83.6&85.4&86.7&88.2\\
		CNN-lc&0.4&71.2&76.1&79.8&81.9&84.2&85.7&87.1&88.7\\
		CNN-lc&0.7&71.3&75.6&78.7&81.6&83.3&85.3&87.0&87.9\\
		CNN-mc&0.0&72.5&75.5&79.0&82.0&84.4&86.3&88.0&89.1\\
		CNN-mc&0.4&72.4&76.1&79.6&82.9&84.7&86.6&88.1&89.6\\
		CNN-mc&0.7&72.5&76.9&79.9&82.9&84.9&86.8&88.2&89.4\\
		CNN-hc&0.0&71.9&75.9&80.1&82.3&85.1&86.8&88.6&89.5\\
		CNN-hc&0.4&72.2&76.3&80.2&83.0&85.0&86.9&88.5&89.8\\
		CNN-hc&0.7&\textbf{73.3}&77.4&80.5&83.2&\textbf{86.5}&87.1&88.8&89.9\\
		ResNet-20&--&62.3&71.4&77.0&80.4&84.1&86.9&\textbf{89.2}&\textbf{90.5}\\
\cellcolor[gray]{0.9}EDLGP(best)&\cellcolor[gray]{0.9}--&\cellcolor[gray]{0.9}73.2&\cellcolor[gray]{0.9}\textbf{79.5}&\cellcolor[gray]{0.9}\textbf{82.1}&\cellcolor[gray]{0.9}\textbf{83.5}&\cellcolor[gray]{0.9}85.0&\cellcolor[gray]{0.9}\textbf{87.2}&\cellcolor[gray]{0.9}88.6&\cellcolor[gray]{0.9}89.0\\
\cellcolor[gray]{0.9}EDLGP(mean)&\cellcolor[gray]{0.9}--&\cellcolor[gray]{0.9}71.8$\pm$1.0&\cellcolor[gray]{0.9}\textbf{78.0$\pm$0.7}&\cellcolor[gray]{0.9}\textbf{81.2$\pm$0.7}&\cellcolor[gray]{0.9}83.0$\pm$0.4&\cellcolor[gray]{0.9}84.5$\pm$0.5&\cellcolor[gray]{0.9}86.1$\pm$0.7&\cellcolor[gray]{0.9}86.8$\pm$0.6&\cellcolor[gray]{0.9}88.0$\pm$0.6\\		
		\hline
		Model&Dropout&sSVHN-10&sSVHN-20&sSVHN-40&sSVHN-80&sSVHN-160&sSVHN-320&sSVHN-640&sSVHN-1280\\
		CNN-lc&0.0&25.3&37.5&50.5&64.1&73.0&77.7&80.9&83.6\\
		CNN-lc&0.4&28.4&44.9&59.0&70.2&77.0&80.3&83.3&85.8\\
		CNN-lc&0.7&28.8&46.7&60.6&72.1&77.7&81.3&83.2&86.2\\
		CNN-mc&0.0&26.8&39.9&53.4&68.6&75.5&79.6&83.2&85.5\\
		CNN-mc&0.4&29.6&43.3&64.3&72.1&78.3&82.2&84.8&87.3\\
		CNN-mc&0.7&27.7&45.8&64.1&74.6&79.7&83.2&86.2&88.2\\
		CNN-hc&0.0&24.9&37.5&55.5&67.7&74.5&80.1&84.0&86.1\\
		CNN-hc&0.4&27.5&45.6&63.1&73.6&79.3&82.7&85.7&88.1\\
		CNN-hc&0.7&28.8&44.8&64.7&74.4&79.5&84.0&86.3&88.6\\
		ResNet-20&--&20.3&40.0&54.7&74.1&\textbf{83.5}&\textbf{86.7}&\textbf{89.5}&\textbf{92.2}\\
	\cellcolor[gray]{0.9}EDLGP(best)&\cellcolor[gray]{0.9}--&\cellcolor[gray]{0.9}\textbf{57.5}&\cellcolor[gray]{0.9}\textbf{66.2}&\cellcolor[gray]{0.9}\textbf{72.8}&\cellcolor[gray]{0.9}\textbf{75.5}&\cellcolor[gray]{0.9}79.7&\cellcolor[gray]{0.9}82.1 &\cellcolor[gray]{0.9}82.1&\cellcolor[gray]{0.9}85.0\\
	\cellcolor[gray]{0.9}EDLGP(mean)&\cellcolor[gray]{0.9}--&\cellcolor[gray]{0.9}\textbf{55.5$\pm$1.5}&\cellcolor[gray]{0.9}\textbf{60.2$\pm$2.4}&\cellcolor[gray]{0.9}\textbf{70.1$\pm$1.5}&\cellcolor[gray]{0.9}73.6$\pm$1.1&\cellcolor[gray]{0.9}76.9$\pm$2.2&\cellcolor[gray]{0.9}79.0$\pm$1.5&\cellcolor[gray]{0.9}80.5$\pm$0.8&\cellcolor[gray]{0.9}82.3$\pm$0.9\\	
		\hline
		\end{tabular}	
		\label{table:testresults}
	\end{center}
	\vspace{-4mm}
\end{table*}

\subsection{Classification Accuracy} 
\textbf{Overall Performance:} In total, there are 40 cases/experiments (i.e., 3$\times$8 on sCIFAR10, sFMNIST and sSVHN, 8 on few-shot sCIFAR10, and 2$\times$4 on ORL and Extended Yale B) to compare the performance EDLGP with different CNNs and existing methods. All the test results are listed in Tables  \ref{table:testresults} - \ref{table:testresultseyale}. In all 40 cases, EDLGP achieves better accuracy in 30 cases among all the existing methods, indicating that EDLGP is effective on these different image classification tasks by automatically evolving variable-length/depth models. Note that the compared methods are the state-of-the-art methods and their results are from the corresponding references. Compared with these methods, EDLGP achieves better performance on the few-shot sCIFAR10, ORL and Extended Yale B datasets in all scenarios with small numbers of training images. On sCIFAR10, sFMNIST and sSVHN, EDLGP achieves better performance when the training set is small, and slightly worse performance when the size of the training set is large. The results show that EDLGP is more effective for data-efficient image classification, particularly the training set is very small.


\begin{table*}[!htbp]
	\footnotesize
	\setlength{\tabcolsep}{0.5em} 
	\caption{Classification accuracy (\%) of EDLGP and benchmark methods of varying complexity on few-shot CIFAR10}
	\vspace{-4mm}
	\begin{center}
		\begin{tabular}{lllllllll}
			\hline 
			Model&sCIFAR10-1&sCIFAR10-2&sCIFAR10-4&sCIFAR10-8&sCIFAR10-16&sCIFAR10-32&sCIFAR10-64&sCIFAR10-128\\
			ResNet \cite{arora2019harnessing}&14.59&17.50&19.52&23.32&28.30&33.15&41.66&49.14\\
			5-layer CNTK \cite{arora2019harnessing}&15.08&18.03&20.83&24.82&29.63&35.26&41.24&47.21\\
			8-layer CNTK \cite{arora2019harnessing}&15.24&18.50&21.07&25.18&30.17&36.05&42.10&48.22\\
			11-layer CNTK \cite{arora2019harnessing}&15.31&18.69&21.23&25.40&30.46&36.44&42.44&48.67\\
			14-layer CNTK \cite{arora2019harnessing}&15.33&18.79&21.34&25.48&30.48&36.57&42.63&48.86\\	
		Harmonic Networks \cite{ulicny2019harmonic222}&11.87&22.24&26.28&\textbf{34.94}&40.47&\textbf{49.59}&\textbf{56.69}&\textbf{63.83}\\	
			\cellcolor[gray]{0.9}EDLGP(best)&\cellcolor[gray]{0.9}\textbf{\textbf{21.10}}&\cellcolor[gray]{0.9}\textbf{23.63}&\cellcolor[gray]{0.9}\textbf{27.42}&\cellcolor[gray]{0.9}{29.93}&\cellcolor[gray]{0.9}\textbf{41.50}&\cellcolor[gray]{0.9}{45.25}&\cellcolor[gray]{0.9}{50.53}&\cellcolor[gray]{0.9}{56.36}\\
			
			\cellcolor[gray]{0.9}EDLGP(mean)&\cellcolor[gray]{0.9}15.22$\pm$2.71&\cellcolor[gray]{0.9}17.69$\pm$3.97&\cellcolor[gray]{0.9}20.63$\pm$4.86&\cellcolor[gray]{0.9}\textbf{25.51$\pm$3.57}&\cellcolor[gray]{0.9}\textbf{37.91$\pm$1.89}&\cellcolor[gray]{0.9}\textbf{41.11$\pm$3.88}&\cellcolor[gray]{0.9}\textbf{47.21$\pm$2.98}&\cellcolor[gray]{0.9}\textbf{54.04$\pm$1.52}\\
			\hline
		\end{tabular}	
		\label{table:testresultsSMALL}
	\end{center}
		\vspace{-6mm}
\end{table*}

\textbf{Results on sCIFAR10, sFMNIST and sSVHN:} The classification accuracies on the test sets are listed in Table \ref{table:testresults}. Note that the test sets are the original one of CIFAR10, Fashion\_MNIST and SVHN and the results of these comparison methods are from \cite{brigato2021close}. On sCIFAR10, EDLGP achieves the best accuracy among all 11 methods including ResNet-20 in 6 cases out of 8 cases, i.e., worse on sCIFAR10 with 1280 training images per class. On very small training data scenarios, EDLGP achieves maximal accuracy that is much higher than the best one of all comparison methods on sCIFAR10. For example, 3.9\% higher (35.8\% vs. 31.9\%) on sCIFAR10-10, 6.2\% higher (43.2\% vs. 37.0\%) on sCIFAR10-20, 5.5\% higher (48.1\% vs. 42.6\%) on sCIFAR10-40, 4.2\% higher (52.3\% vs. 48.1\%) on sCIFAR10-80, and 4.3\% higher (58.2\% vs. 53.9\%) on sCIFAR10-160. The results further demonstrate that EDLGP is very effective when the training set is small. On Fashion\_MNIST, EDLGP achieves the best results on sFMNIST-20, sFMNIST-40, sFMNIST-80, and sFMNIST-640. In the remaining cases, EDLGP achieves slightly worse accuracy than the best accuracy, i.e., 0.1\% gap on sFMNIST-10, 1.5\% gap on sFMNIST-160, 0.6\% gap on sFMNIST-640, and 1.5\% gap on sFMNIST-1280. The results show that EDLGP is effective for classifying the sFMNIST dataset. On sSVHN, EDLGP achieves the best results on sSVHN-10, sSVHN-20, sSVHN-40, and sSVHN-80. More importantly, the best accuracy obtained by EDLGP is much higher than that by all comparison methods, i.e., 27.9\% higher on sSVHN-10, 19.5\% higher on sSVHN-20, and 8.1\% on sSVHN-40. With the increasing of training data, some CNNs and ResNet-20 methods tend to achieve better classification accuracy and the accuracy gap between EDLGP and the best CNN become smaller. On sSVHN-160, sSVHN-320, sSVHN-640, and sSVHN-1280, EDLGP achieves better results than the majority of the comparison methods and worse results than ResNet-20. The results show that CNNs require sufficient training data to obtain good classification performance. Compared with these CNN methods, EDLGP is significantly less data intensive and can achieve better performance when the training set is very small. CNN models typically contain a huge number of parameters so that a large number of training images are needed to train. Unlike CNNs, EDLGP automatically evolves tree-based models of functions and operators that have significantly fewer parameters and does not require a large number of training images. EDLGP is more data efficient than CNNs since it can achieve better performance than CNNs when the training data is small.

\textbf{Results on few-shot sCIFAR10:} Table \ref{table:testresultsSMALL} shows the results on few-shot sCIFAR10 with 1, 2, 4, 8, 16, 32, 64, 128 training images per class, respectively. Among all the methods including ResNet, EDLGP achieves the best classification accuracy on few-shot sCIFAR10 with 1-128 training images per class. The best accuracy obtained by EDLGP is much higher than the best accuracy of all the comparison methods, i.e., over 7\% on average. In these few-shot settings, both the performance of CNN and EDLGP increase with the number of training images, but EDLGP performs better than these CNN methods. The results continually show the effectiveness of EDLGP in data-efficient image classification.

\begin{table}[!htbp]
		\vspace{-4mm}
			\footnotesize
	\setlength{\tabcolsep}{0.7em} 
	\caption{Classification accuracy (\%) of EDLGP and the benchmark methods on the ORL dataset}
		\vspace{-4mm}
	\begin{center}
		\begin{tabular}{lllll}
			\hline 
		Method&ORL-2&ORL-3&ORL-4&ORL-5\\ \hline
		Eigenface \cite{turk1991eigenfaces} &70.7$\pm$2.7&78.9$\pm$2.3&84.2$\pm$2.1&87.9$\pm$2.5\\
		Fisherface \cite{belhumeur1997eigenfaces}&75.5$\pm$3.3&86.1$\pm$1.9&91.6$\pm$1.9&94.3$\pm$1.4\\
		Laplacianface \cite{he2005face}&77.6$\pm$2.5&86.0$\pm$2.0&90.3$\pm$1.7&93.0$\pm$1.9\\
		NPE \cite{he2005neighborhood}&77.6$\pm$2.7&85.7$\pm$1.8&90.5$\pm$1.8&93.4$\pm$1.8\\
		MFA \cite{yan2006graph}&75.4$\pm$3.1&86.1$\pm$1.9&91.6$\pm$1.9&94.3$\pm$1.4	\\		
		CNN \cite{liao2021deep}&69.7$\pm$3.1&82.9$\pm$2.5&87.9$\pm$1.8&91.5$\pm$2.9\\	
		KCNN \cite{liao2021deep}&72.8$\pm$3.1&84.6$\pm$2.6&91.7$\pm$2.4&94.6$\pm$1.5\\
		\cellcolor[gray]{0.9}EDLGP(best)&\cellcolor[gray]{0.9}\textbf{90.3}&\cellcolor[gray]{0.9}\textbf{97.9}&\cellcolor[gray]{0.9}\textbf{99.6}&\cellcolor[gray]{0.9}\textbf{99.5}\\	
		\cellcolor[gray]{0.9}EDLGP(mean)&\cellcolor[gray]{0.9}\textbf{86.4$\pm$3.9}&\cellcolor[gray]{0.9}\textbf{96.4$\pm$1.0}&\cellcolor[gray]{0.9}\textbf{98.4$\pm$1.1}&\cellcolor[gray]{0.9}\textbf{99.3$\pm$0.2}\\
			\hline
		\end{tabular}	
		\label{table:testresultsatt}
	\end{center}
		\vspace{-2mm}
\end{table}

\textbf{Results on ORL and Extended Yale B:} On the face datasets, EDLGP achieves higher maximum and average accuracies than any of the comparison methods in all scenarios. Specifically, EDLGP improves the average accuracy by 8.8\% on ORL-2, 10.3\% on ORL-3, 6.7\% on ORL-4, and 4.7\% on ORL-5. On ORL-5, EDLGP achieves an average accuracy of 99.3\% and maximum accuracy of 99.5\%. Furthermore, compared with these seven methods, EDLGP has a smaller standard deviation value, which means EDLGP is more stable. On Extended Yale B, EDLGP gains over 99\% accuracy in all scenarios. Among all the comparison methods, EDLGP achieves the best accuracy on Extended Yale B, which means that EDLGP is more accurate than the 12 different comparison methods. To sum up, EDLGP is effective for face image classification using few training images.

\begin{table}
 \vspace{-6mm}
	\setlength{\tabcolsep}{0.45em} 
	\caption{Classification accuracy (\%) of EDLGP and the benchmark methods on the Extended Yale B dataset. 15, 20, 25, and 30 denote different numbers of training images in each class}
		\vspace{-4mm}
	\begin{center}
		\begin{tabular}{lllll}
			\hline 
			Method&15&20&25&30\\ \hline
			CRC \cite{zhang2011sparse}&91.39&94.26&95.91&97.04\\
			SRC \cite{wright2008robust}&91.72&93.71&95.56&96.37\\
			CESR \cite{he2010maximum}&77.92&83.42&85.68&88.51\\
			RSC \cite{yang2011robust}&95.01&97.04&97.81&98.40\\
			HQA \cite{he2013half}&93.39&93.99&90.19&92.41\\
			HQM \cite{he2013half}&91.14&94.15&95.29&96.46\\
			NMR \cite{yang2016nuclear}&93.50&96.29&97.57&98.54\\
			RMR \cite{xie2017robust}&93.56&94.08&92.15&92.72\\
			FDDL \cite{yang2014sparse}&93.44&94.92&96.38&96.94\\
			LRSDL \cite{vu2017fast}&94.92&96.69&97.88&98.31\\
			DCM(N) \cite{zhang2019deep}&93.17&95.97&97.38&98.38\\
			DCM(S) \cite{zhang2019deep}&98.87&99.51&99.63&99.79\\
			\cellcolor[gray]{0.9}EDLGP(best)&\cellcolor[gray]{0.9}\textbf{99.89}&\cellcolor[gray]{0.9}\textbf{99.94}&\cellcolor[gray]{0.9}\textbf{99.93}&\cellcolor[gray]{0.9}\textbf{99.84}\\	
			\cellcolor[gray]{0.9}EDLGP(mean)&\cellcolor[gray]{0.9}\textbf{99.80$\pm$0.05}&\cellcolor[gray]{0.9}\textbf{99.80$\pm$0.09}&\cellcolor[gray]{0.9}\textbf{99.83$\pm$0.08}&\cellcolor[gray]{0.9}\textbf{99.82$\pm$0.04}\\
			\hline
		\end{tabular}	
		\label{table:testresultseyale}
	\end{center}
		\vspace{-2mm}
\end{table}

To summarise, EDLGP achieves promising performance in image classification with small training data. Unlike comparative methods, EDLGP can simultaneously and automatically search for the best feature extraction methods and ensemble models to achieve effective classification. The learned EDLGP models have fewer parameters than CNNs so that they can be better trained on small training data and achieve higher classification accuracy. A large number of comparisons show that EDLGP is significantly more data efficient than CNNs and other well-known methods on different types of image classification tasks. The results comprehensively verify the effectiveness and superiority of EDLGP in data-efficient image classification.

 \subsection{Running and Classification Time}
\begin{figure}
 	\centering
 	\includegraphics[width=\linewidth]{ 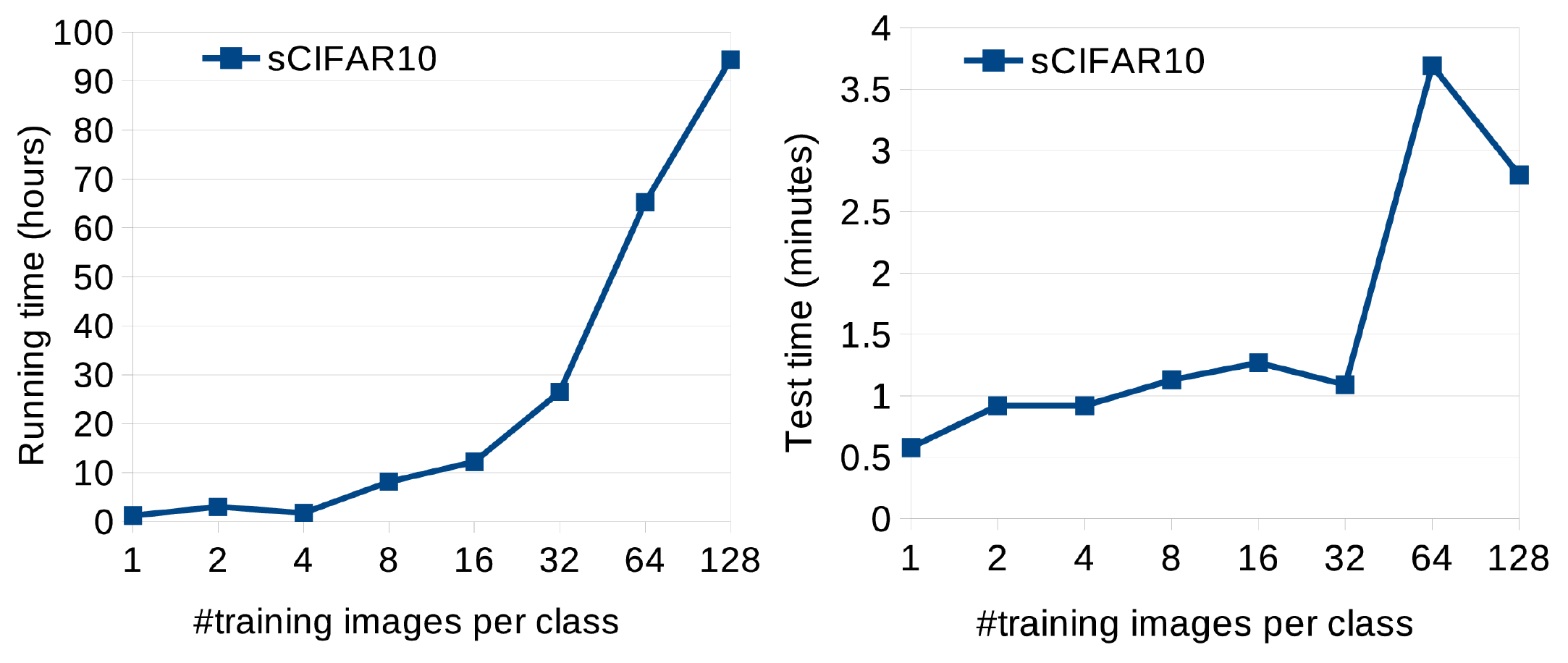}
 	\vspace{-6mm}
 	\caption{Running time and test/classification time of EDLGP on the sCIFAR10 dataset using 1-128 images/instances per class for training.}
 	\label{fig:running_time_scifar10_2}
 	\vspace{-4mm}
 \end{figure}

 We take the few-shot sCIFAR10-X (X ranges from 1 to 128) dataset as a typical example to analyse the running and classification time of EDLGP. The running time of EDLGP on the other datasets is analysed in the supplementary materials due to the page limit. Figure \ref{fig:running_time_scifar10_2} shows an average running time and classification time of EDLGP on a single CPU. The running time of EDLGP is less than 10 hours when the number of training images per class is smaller than 16. The running time of EDLGP increases gradually with the number of training images. On sCIFAR10-128, it uses more than 90 hours to complete the evolutionary learning process. The running time of EDLGP is clearly longer than that of CNNs, although we did not directly compare them. The main reason is that EDLGP is a population-based search algorithm that is currently implemented using the DEAP package running on CPUs. In contrast, CNNs can benefit from a fast computing facility--GPU, and use a shorter time for model training. The running time of EDLGP can be accelerated by using a GPU implementation or running it in parallel on multiple CPUs.

 The classification/test time of EDLGP on sCIFAR10-X is fast, i.e., less than 4 minutes in all scenarios. In the classification process, EDLGP trains the model found via evolution to classify 10,000 images. The GP model complexity is the main factor that affects the classification time. From Fig. \ref{fig:running_time_scifar10_2}, EDLGP may learn a more complex model on sCIFAR10-64 and sCIFAR10-128 than the other scenarios, as it uses a longer classification time. The analysis of the model length/complexity will be conducted in the following subsection. Overall, EDLGP needs a reasonably long running time to complete the evolutionary process but uses a short time for classification. 


\subsection{Convergence Behaviours}
\begin{figure}
	\centering
	\includegraphics[width=\linewidth]{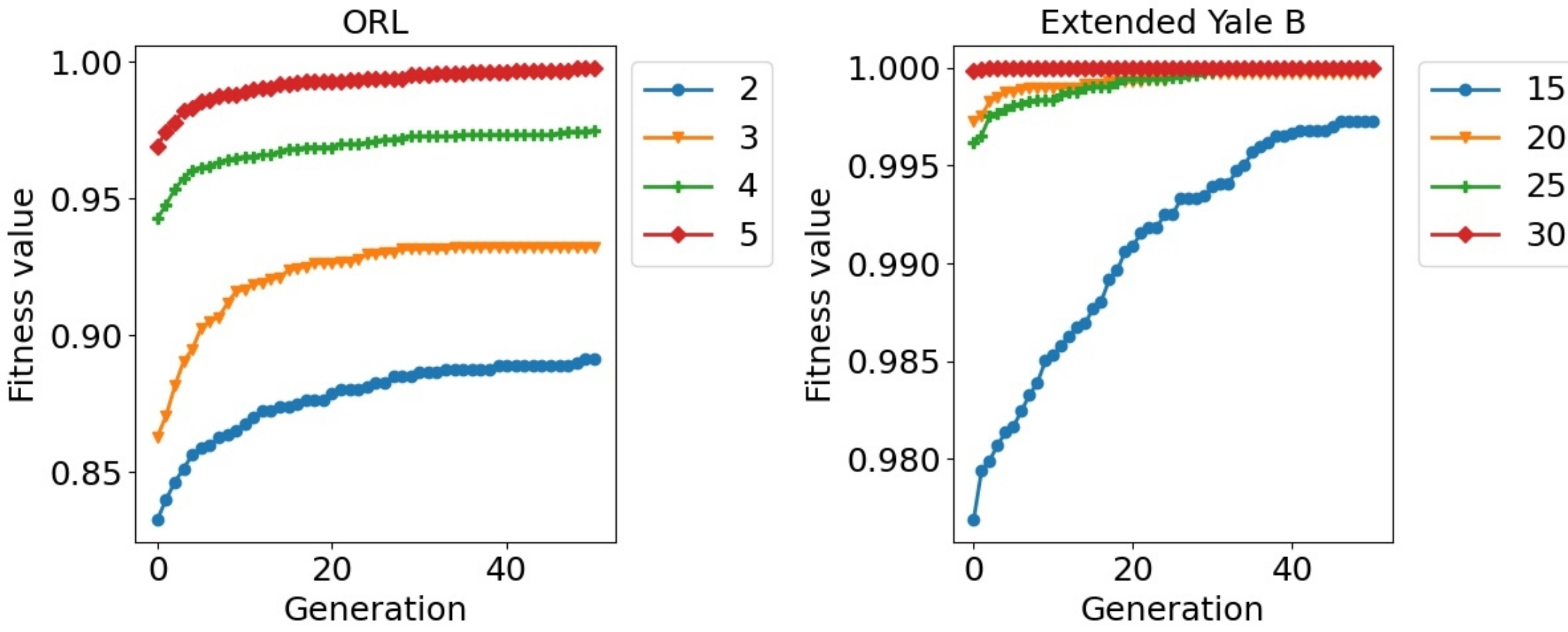}
		\vspace{-6mm}
	\caption{Convergence behaviours of EDLGP on ORL and Extended Yale B.}
	\label{fig:convergence_att}
		\vspace{-4mm}
\end{figure}

We take the ORL and Extended Yale B datasets as examples to show the convergence behaviour of EDLGP. Note that its convergence behaviour on other datasets shows similar patterns. Figure \ref{fig:convergence_att} shows that EDLGP has good search ability and can converge to a high fitness value (i.e. accuracy on the training set) after 50 generations. Using different training set sizes, EDLGP achieves different fitness values during evolution. Specifically, more training data corresponds to higher fitness values of EDLGP on these two datasets. This is because the fitness function evaluates the generalisation of the learned model by using k-fold cross-validation on the training set. To sum up, EDLGP has good search ability and convergence, and can find the best model through evolution. 

\subsection{Interpretability of Trees/Models}
This subsection analyses model length/complexity and visualisation to demonstrate the high interpretability of models learned by EDLGP. 

\begin{figure}
	\centering
	\includegraphics[width=\linewidth]{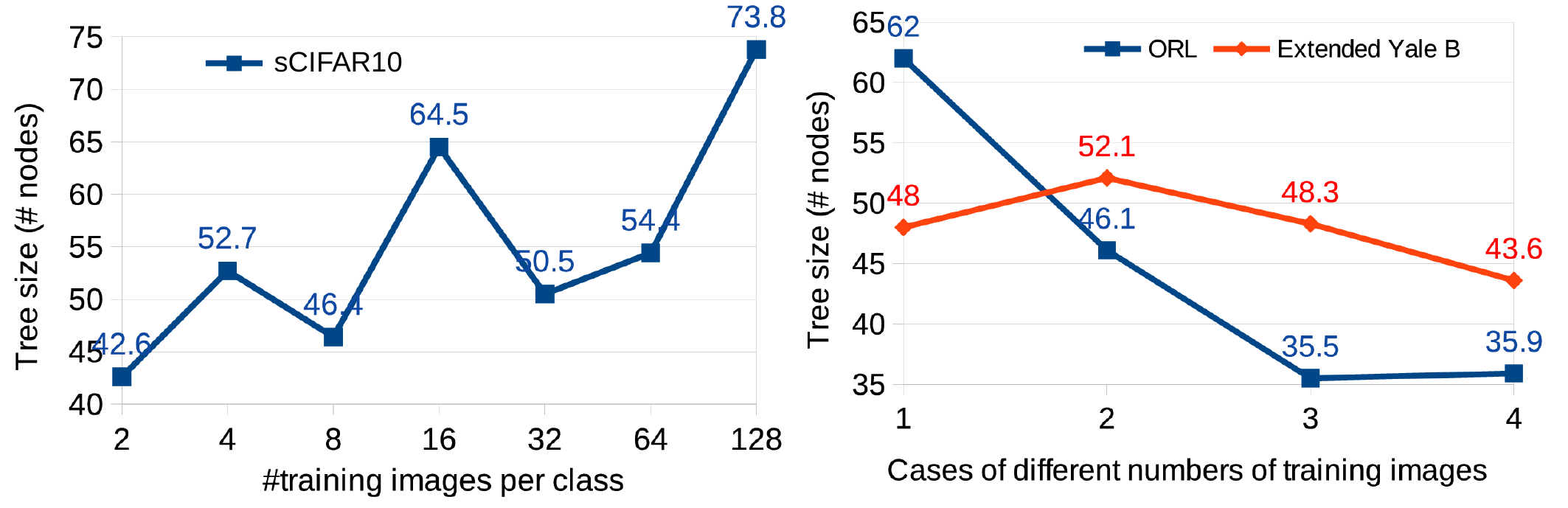}
	\vspace{-4mm}
	\caption{The average size (number of nodes) of trees evolved by EDLGP on sCIFAR10, ORL, and Extended Yale B of different training set sizes.}
	\label{fig:tree_size_cifar_att}
	\vspace{-4mm}
\end{figure}

\textbf{Tree/Model Size:} The average tree size of EDLGP on sCIFAR10, ORL and Extended Yale B is shown in Fig. \ref{fig:tree_size_cifar_att}. On sCIFAR10 with 1-128 training images, the average tree size ranges from 42.6 to 73.8. EDLGP tends to gradually increase its tree size with the number of training images on sCIFAR10, which is a difficult task. On the ORL and Extended Yale B datasets, the average tree size ranges from 35.5 to 62. In case 1 (2 training images on ORL and 15 training images on Extended Yale B), EDLGP has a small initial fitness value (as shown in Fig. \ref{fig:convergence_att}) and seems to improve tree quality by gradually increasing the tree size. When using more training images on these two datasets (i.e. in cases 3 and 4), EDLGP reaches a higher fitness value in initial generations and finds relatively smaller trees than that in case 1. To sum up, EDLGP can evolve variable-length trees on different datasets with various numbers of training images.

\begin{figure}
	\centering
	\includegraphics[width=\linewidth]{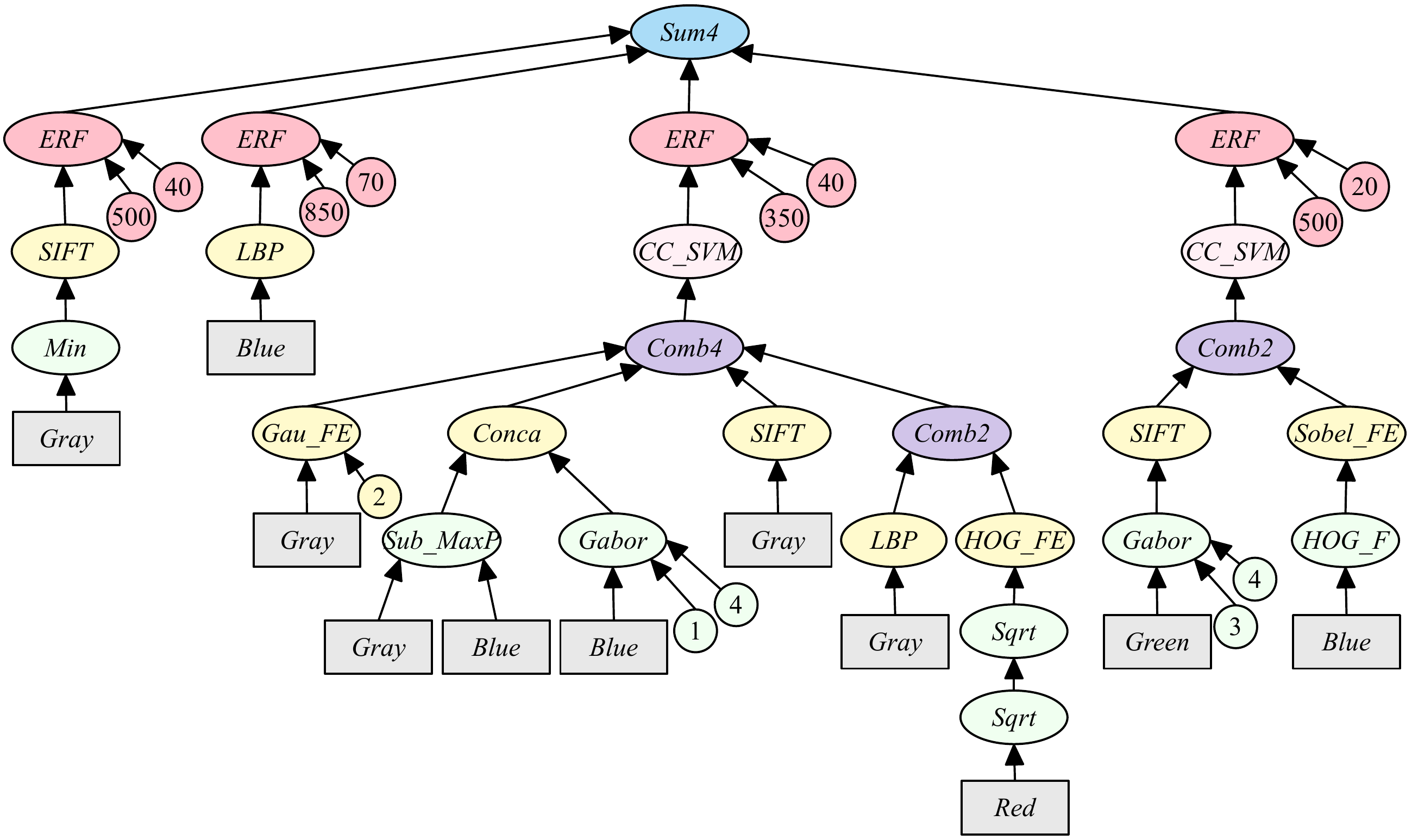}
	\vspace{-6mm}
	\caption{An example tree of EDLGP on sCIFAR10 using 80 training images per class. This tree achieves 52.26\% accuracy on the test set, which is better than any of the comparison methods.}
	\label{fig:example_tree_cifar}
	\vspace{-4mm}
\end{figure}

\textbf{Tree/Model Visualisation:} Figure \ref{fig:example_tree_cifar} shows an example tree of EDLGP on sCIFAR10-80. This tree is an ensemble of four ERF ensemble classifiers. Each branch can build one ERF classifier using different features generated from the corresponding child nodes and different parameter settings (i.e. number of decision trees and maximal tree depth). It processes images using Min, Sub\_MaxP, Gabor, Sqrt, and HOG\_F operators on the input image in the gray, red, blue, and green channels, and extracts features using SIFT, LBP, Gau\_FE, Conca, HOG\_FE, and Sobel\_FE operators. CIFAR10 is a complex object classification dataset so that different types of features are extracted to improve classification performance. By using such a model, EDLGP achieves 52.26\% test accuracy, which is the best accuracy among all the methods on sCIFAR10 using only 80 training images per class. More importantly, the size of this tree is much smaller than those CNNs. 

\begin{figure}
	\centering
	\includegraphics[width=\linewidth]{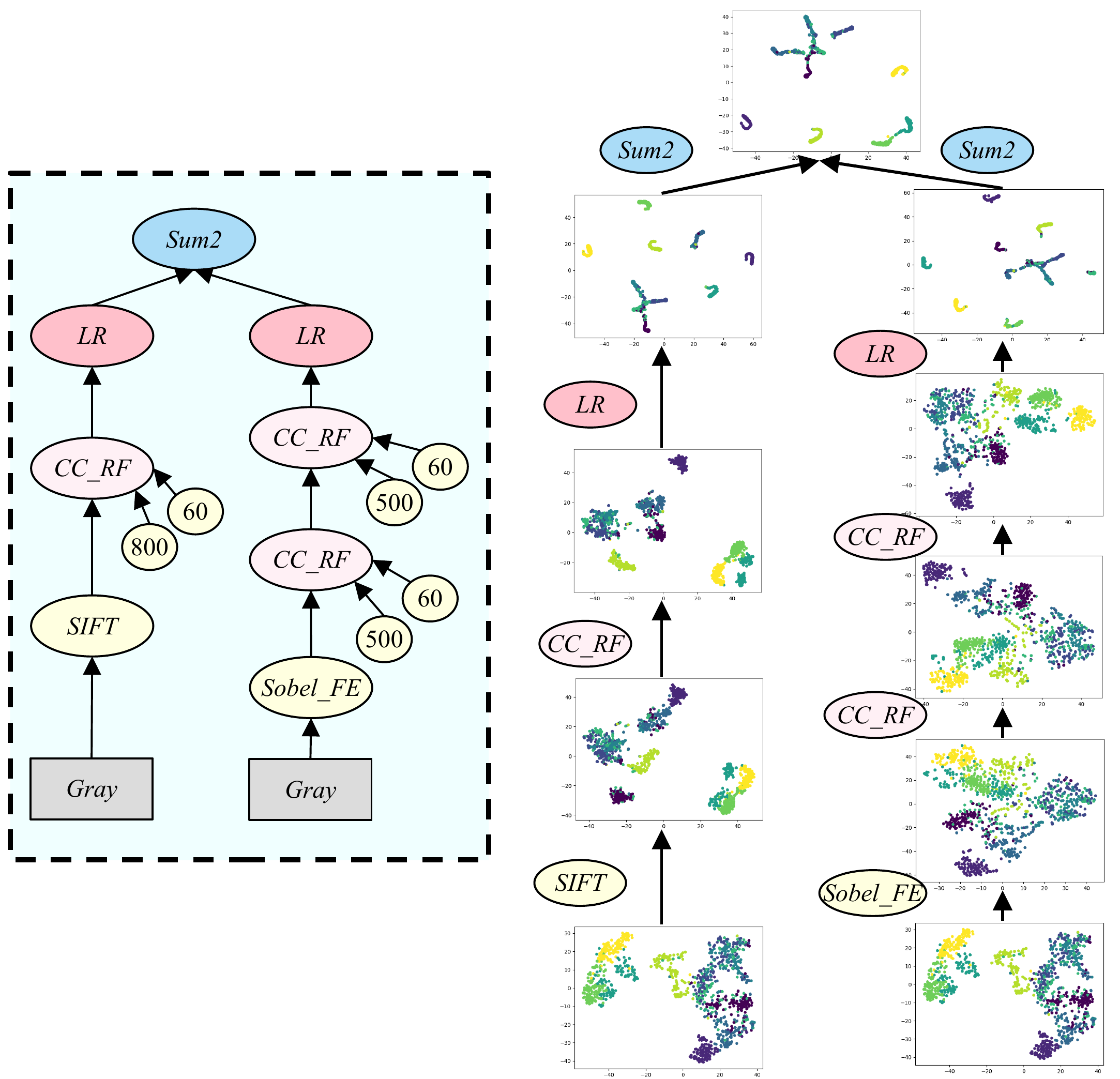}
	\vspace{-6mm}
	\caption{An example tree evolved by EDLGP on sFMNIST-640 and visualisation of the test data transformed by each internal node of the tree. T-SNE \cite{maaten2008visualizing} is used for visualisation. Each plot is drawn using 100 testing images per class of sFMNIST and each colour represents one class.}
	\label{fig:exampletreefmnist640}
	\vspace{-4mm}
\end{figure}

To further demonstrate the interpretability of models/trees evolved by EDLGP, a small tree on sFMNIST-640 is analysed. The small tree is shown in Fig. \ref{fig:exampletreefmnist640}. It achieves 86.7\% test accuracy on sFMNIST-640. We use T-SNE \cite{maaten2008visualizing} to visualise the data (i.e., features) generated by each internal node of the GP tree. For better visualisation, 100 test images per class are used and each test image is fed into the GP tree. We collect all the outputs of each internal node by feeding these test images and use T-SNE to perform visualisation on the outputs, as shown in Fig. \ref{fig:exampletreefmnist640}. This example tree has two branches to generate an ensemble of LR classifiers. For the left branch, it extracts SIFT features, uses CC\_RF to generate a new feature vector, and performs classification using LR. The right branch extracts features using Sobel\_FE, generates new features using two CC\_RF functions, and performs classification using LR. The visualised data in Fig. \ref{fig:exampletreefmnist640} shows that the function at each node makes significant changes to the input images/features. The data generated by the high-level nodes are more clustered. At the output node, the data generated from the Sum2 node are clearly clustered, which can reveal why this model can achieve high classification accuracy.  


To sum up, EDLGP evolves trees with different lengths, shapes and depths on different datasets. A single GP tree is a model that can perform image feature extraction and classification using ensembles of classifiers. The functions/internal nodes of GP trees can make important transformations on the data to generate better ones for achieving good performance. The GP trees are easy to be visualised and some insights can be gained from them. 

\subsection{Transferability of Trees/Models}
This subsection analyses the transferability of trees/models found by EDLGP to show the possibility of model reuse. The best EDLGP models found from sCIFAR10 using 1-128 training images per class is transferred to solve sSVNH using 10-1280 training images per class. The average accuracy (\%) is presented in Table \ref{table:cifar10toSvHN}. The evolved EDLGP models are also transferred between the ORL and Extended Yale B datasets of different training set sizes. The average accuracy (\%) is presented in Table \ref{table:transfer_orl_eyaleb}. Note that in the transferring setting the classifiers of the GP trees learned on the source training set are re-trained using the training set of the target task and the GP trees are used to classify the test set of the target task. 

\begin{table}[!htbp]
\vspace{-4mm}
	\footnotesize
	\setlength{\tabcolsep}{0.4em} 
	\caption{Average classification accuracy (\%) on sSVHN using the models learned by EDLGP from sCIFAR10-1 to sCIFAR10-128.}
		\vspace{-2mm}
	\begin{center}
 \begin{tabular}{lllllllll}
    \hline
        \multicolumn{9}{c}{sCIFAR10 $\longrightarrow$ sSVHN}\\\hline
    &10&20&40&80&160 &320 &640 &1280 \\ \hline
sCIFAR10-1 & 28.2 & 32.3 & 37.0 & 42.3 & 47.4 & 52.0 & 56.0 & 59.2 \\
sCIFAR10-2 & 23.3 & 26.9 & 29.5 & 32.2 & 34.1 & 35.9 & 37.5 & 38.8 \\ 
sCIFAR10-4 & 34.8 & 41.6 & 46.7 & 51.6 & 55.8 & 59.1 & 61.5 & 62.8 \\ 
sCIFAR10-8 & 27.9 & 31.9 & 36.3 & 39.4 & 42.8 & 45.9 & 48.5 & 50.9 \\
sCIFAR10-16 & 40.6 & 47.5 & 53.0 & 58.9 & 63.0 & 67.0 & 70.0 & 71.8 \\ 
sCIFAR10-32 & 46.8 & 54.4 & 60.0 & 65.0 & 68.1 & 71.3 & 73.5 & 75.1 \\ 
sCIFAR10-64 & \textbf{52.0 }& \textbf{59.4} & \textbf{63.8} & \textbf{68.4} & \textbf{71.5} & 74.6 & 76.5 & 78.2 \\ 
sCIFAR10-128 & 47.2 & 55.5 & 61.4 & 67.6 & 71.4 & \textbf{75.2} & \textbf{77.5} & \textbf{79.1} \\ 
\hline
ResNet-20(original)&20.3&40.0&54.7&74.1&83.5&86.7&89.5&92.2\\
EDLGP(original)&55.5&60.2&70.1&73.6&76.9&79.0&80.5&82.3\\	
    \hline
    \end{tabular}
		\label{table:cifar10toSvHN}
	\end{center}
		\vspace{-4mm}
\end{table}

Table \ref{table:cifar10toSvHN} lists the classification results on sSVHN (target dataset) by reusing the best trees/models evolved by EDLGP on few-shot sCIFAR10 (source dataset). The final two rows list the original results of ResNet-20 and EDLGP learning from the target training set of sSVHN for comparisons. The results show that the learned EDLGP models have very strong transferability since they achieve better classification performance than ResNet-20 on sSVHN-10, sSVHN-20 and sSVHN-40 and very close accuracy to EDLGP in all cases. Comparing with other CNN methods in Table \ref{table:testresults}, the reused models can also obtain higher accuracy in most cases. The comparison can better demonstrate good transferability of the EDLGP models. From Table \ref{table:testresults}, we can also find that the transferability of the EDLGP models is also affected by the number of training instances in the source training set. Specifically, a relatively larger source training set can lead to better transferability of the EDLGP models on sSVHN. This is reasonable because both CIFAR10 and SVHN are object classification tasks and more data can lead to learn more generalised models. However, EDLGP does not heavily rely on increasing the the training set to improve its performance and the model transferability, since the models evolved on sCIFAR10-64 achieve the best performance on sSVHN-10, sSVHN-20, sSVHN-40, and sSVHN-80.

Table \ref{table:transfer_orl_eyaleb} shows the results obtained by transferring EDLGP models across the ORL and Extended Yale B datasets. The final row lists the results obtained by EDLGP using the target training set for model learning. From ORL to Extended Yale B, the reused EDLGP models achieve worse accuracy when using 15 training images per class, but very high accuracy (i.e. 88.6\%) when using 30 training images per class. From Extended Yale B to ORL, the reused GP models obtain better accuracy than some comparison methods on ORL with different numbers of training images, e.g., better than Eigenface on all cases, better than CNN on ORL-2, ORL-3 and ORL4. The results can show the high transferability of the learned GP models across different face image classification tasks. Table \ref{table:transfer_orl_eyaleb} shows that using different training set sizes in the source and target tasks, the models achieve different performances. In the case of ORL$\longrightarrow$Extended Yale B, models learned from a small training set of ORL can achieve better performance on Extended Yale B. In the case of Extended Yale B$\longrightarrow$ORL, models learned from 20 source training images per class achieve better performance on ORL with different training set sizes. The two face image datasets are different in terms of image variations, which may affect the generality and transferability of the models learned by EDLGP. 

To sum up, the results show that the models/trees learned by EDLGP have high transferability. An important reason is that the learned models/trees consisting of operators and algorithms can represent not only domain-specific information but also general-domain information, which enable them to have strong transferability. This reveals the high possibility of reusing the learned EDLGP models for solving other similar or related tasks. In addition, the results show that the source training set sizes may affect the transferability of the EDLGP models, which provides insights on effective model reuse. 

\begin{table}[!htbp]
\vspace{-4mm}
	\setlength{\tabcolsep}{0.4em} 
	\caption{Average test accuracy (\%) by transferring GP models between the ORL and Extended Yale B datasets using different numbers of training images}
		\vspace{-4mm}
	\begin{center}
    \begin{tabular}{lllll|lllll}
    \hline
    \multicolumn{5}{c|}{ORL $\longrightarrow$ Extended Yale B}&\multicolumn{5}{c}{Extended Yale B $\longrightarrow$ ORL}\\\hline
        ~ & 15 & 20 & 25 & 30 & ~ & 2 & 3 & 4 & 5 \\ \hline
        ORL-2 & \textbf{76.7} &\textbf{ 81.8} & \textbf{85.9} & \textbf{88.6} & Extended.-15 & 62.3 & 73.8 & 80.3 & 83.2 \\ 
        ORL-3 & 70.5 & 76.7 & 80.9 & 85.6 & Extended.-20& \textbf{71.2} & \textbf{83.8} & \textbf{89.3} & 91.5 \\ 
        ORL-4 & 68.7 & 74.0 & 77.9 & 81.5 & Extended.-25& 55.5 & 68.1 & 76.2 & 81.7 \\ 
        ORL-5 & 69.4 & 76.2 & 80.3 & 84.2 & Extended.-30 & 69.6 & 82.6 & 88.0 & \textbf{91.8} \\ \hline
 	   EDLGP&99.80&99.8&99.8&99.8&EDLGP&86.4&96.4&98.4&99.3\\\hline
    \end{tabular}
		\label{table:transfer_orl_eyaleb}
	\end{center}
		\vspace{-6mm}
\end{table}



\subsection{Summary}
The following observations of EDLGP as a typical example of GP-based EDL method can be summarised from the experimental analysis. 

\begin{itemize}
    \item EDLGP can achieve better performance than the compared methods in most cases on CIFAR10, Fashion\_MNIST, SVHN, ORL, and Extended Yale B. When the training set is very small, EDLGP can achieve state-of-the-art performance. This shows that EDLGP is an effective approach to data-efficient image classification. 
    
    
    \item EDLGP shows good convergence and can find trees with different shapes, sizes and depths on different detasets. Compared with CNN-based image classification methods, EDLGP does not require manually tune/design/determine the model architectures and/or coefficients/parameters. 
    
    \item EDLGP can evolve small and easy-interpretable trees to achieve high accuracy. The evolved EDLGP trees are composed of image and classification domain operators, which are interpretable to provide more insights into the tasks. This is difficult to be achieved by using CNN methods with numerous parameters. 
    
    \item The trees/models evolved by EDLGP show high transferability, facilitating model reuse and development in the future.
\end{itemize}

\section{Conclusions}
In this paper, a GP-based EDL method, EDLGP, was proposed for automatically evolving variable-length models for image classification on small training data. A new representation was developed that includes a multi-layer tree structure, a function set and a terminal set, enabling EDLGP to efficiently build models to perform feature extraction, concatenation, classification and cascade, and ensemble construction, automatically and simultaneously. The EDLGP approach has shown great potential in solving data-efficient image classification by achieving better performance than many effective methods. A detailed analysis showed that the models learned by EDLGP have good convergence, high interpretability, and good transferability. Compared with existing popular CNN-based image classification methods, the EDLGP approach as a GP-based EDL approach has a number of advantages, such as without requiring expensive GPU devices to run and rich domain expertise to design/tune model architectures/coefficients, data-efficient, and evolving variable-length models with high interpretability and transferability.

This paper is a starting point showing the superiority of GP in comparisons with CNN-based methods for data-efficient image classification. More effective GP-based EDL methods can be developed to dig out the potential of GP in the future. 


\bibliographystyle{IEEEtran}
\bibliography{ref.bib}

\begin{thebibliography}{77}
\providecommand{\natexlab}[1]{#1}
\providecommand{\url}[1]{#1}
\csname url@samestyle\endcsname
\providecommand{\newblock}{\relax}
\providecommand{\bibinfo}[2]{#2}
\providecommand{\BIBentrySTDinterwordspacing}{\spaceskip=0pt\relax}
\providecommand{\BIBentryALTinterwordstretchfactor}{4}
\providecommand{\BIBentryALTinterwordspacing}{\spaceskip=\fontdimen2\font plus
\BIBentryALTinterwordstretchfactor\fontdimen3\font minus
  \fontdimen4\font\relax}
\providecommand{\BIBforeignlanguage}[2]{{%
\expandafter\ifx\csname l@#1\endcsname\relax
\typeout{** WARNING: IEEEtranN.bst: No hyphenation pattern has been}%
\typeout{** loaded for the language `#1'. Using the pattern for}%
\typeout{** the default language instead.}%
\else
\language=\csname l@#1\endcsname
\fi
#2}}
\providecommand{\BIBdecl}{\relax}
\BIBdecl

\bibitem[Schindelin and et~al.(2012)]{schindelin2012fiji}
J.~Schindelin and et~al., ``Fiji: an open-source platform for biological-image
  analysis,'' \emph{Nature methods}, vol.~9, no.~7, pp. 676--682, 2012.

\bibitem[Schneider et~al.(2012)Schneider, Rasband, and
  Eliceiri]{schneider2012nih}
C.~A. Schneider, W.~S. Rasband, and K.~W. Eliceiri, ``Nih image to imagej: 25
  years of image analysis,'' \emph{Nature methods}, vol.~9, no.~7, pp.
  671--675, 2012.

\bibitem[Lu and Weng(2007)]{lu2007survey}
D.~Lu and Q.~Weng, ``A survey of image classification methods and techniques
  for improving classification performance,'' \emph{Int. J. Remote Sens.},
  vol.~28, no.~5, pp. 823--870, 2007.

\bibitem[Bi et~al.(2021{\natexlab{a}})Bi, Xue, and Zhang]{bi2021gpimage}
Y.~Bi, B.~Xue, and M.~Zhang, \emph{Genetic Programming for Image
  Classification: An Automated Approach to Feature Learning}.\hskip 1em plus
  0.5em minus 0.4em\relax Springer, 2021.

\bibitem[LeCun et~al.(2015)LeCun, Bengio, and Hinton]{lecun2015deep}
Y.~LeCun, Y.~Bengio, and G.~Hinton, ``Deep learning,'' \emph{nature}, vol. 521,
  no. 7553, pp. 436--444, 2015.

\bibitem[{Bengio} et~al.(2013){Bengio}, {Courville}, and
  {Vincent}]{bengio2013representationlearning}
Y.~{Bengio}, A.~{Courville}, and P.~{Vincent}, ``Representation learning: A
  review and new perspectives,'' \emph{IEEE Trans. Pattern Anal. Mach.
  Intell.}, vol.~35, no.~8, pp. 1798--1828, 2013.

\bibitem[Zhou and Feng(2018)]{zhou2018deep}
Z.-H. Zhou and J.~Feng, ``Deep forest,'' \emph{Natl. Sci. Rev.}, vol.~6, no.~1,
  pp. 74--86, Oct 2018.

\bibitem[Alzubaidi et~al.(2021)Alzubaidi, Zhang, Humaidi, Al-Dujaili, Duan,
  Al-Shamma, Santamar{\'\i}a, Fadhel, Al-Amidie, and
  Farhan]{alzubaidi2021review}
L.~Alzubaidi, J.~Zhang, A.~J. Humaidi, A.~Al-Dujaili, Y.~Duan, O.~Al-Shamma,
  J.~Santamar{\'\i}a, M.~A. Fadhel, M.~Al-Amidie, and L.~Farhan, ``Review of
  deep learning: Concepts, cnn architectures, challenges, applications, future
  directions,'' \emph{J. Big Data}, vol.~8, no.~1, pp. 1--74, 2021.

\bibitem[Chan et~al.(2015)Chan, Jia, Gao, Lu, Zeng, and Ma]{chan2015pcanet}
T.-H. Chan, K.~Jia, S.~Gao, J.~Lu, Z.~Zeng, and Y.~Ma, ``Pcanet: A simple deep
  learning baseline for image classification?'' \emph{IEEE Trans. Image
  Process.}, vol.~24, no.~12, pp. 5017--5032, 2015.

\bibitem[Zhan et~al.(2022)Zhan, Li, and Zhang]{zhan2022evolutionary}
Z.-H. Zhan, J.-Y. Li, and J.~Zhang, ``Evolutionary deep learning: A survey,''
  \emph{Neurocomputing}, 2022.

\bibitem[Al-Sahaf et~al.(2019)Al-Sahaf, Bi, Chen, Lensen, Mei, Sun, Tran, Xue,
  and Zhang]{al2019survey}
H.~Al-Sahaf, Y.~Bi, Q.~Chen, A.~Lensen, Y.~Mei, Y.~Sun, B.~Tran, B.~Xue, and
  M.~Zhang, ``A survey on evolutionary machine learning,'' \emph{J. Roy. Soc.
  New Zeal.}, vol.~49, no.~2, pp. 205--228, 2019.

\bibitem[B{\"a}ck et~al.(1997)B{\"a}ck, Fogel, and
  Michalewicz]{back1997handbook}
T.~B{\"a}ck, D.~B. Fogel, and Z.~Michalewicz, ``Handbook of evolutionary
  computation,'' \emph{Release}, vol.~97, no.~1, p.~B1, 1997.

\bibitem[Zhou et~al.(2021)Zhou, Qin, Gong, and Tan]{zhou2021survey}
X.~Zhou, A.~Qin, M.~Gong, and K.~C. Tan, ``A survey on evolutionary
  construction of deep neural networks,'' \emph{IEEE Trans. Evol. Comput.},
  2021.

\bibitem[Liu et~al.(2021, DOI: 10.1109/TNNLS.2021.3100554)Liu, Sun, Xue, Zhang,
  Yen, and Tan]{liu2020survey}
Y.~Liu, Y.~Sun, B.~Xue, M.~Zhang, G.~G. Yen, and K.~C. Tan, ``A survey on
  evolutionary neural architecture search,'' \emph{IEEE Trans. Neural Netw.
  Learn. Syst.}, 2021, DOI: 10.1109/TNNLS.2021.3100554.

\bibitem[Martinez et~al.(2021)Martinez, Del~Ser, Villar-Rodriguez, Osaba,
  Poyatos, Tabik, Molina, and Herrera]{martinez2021lights}
A.~D. Martinez, J.~Del~Ser, E.~Villar-Rodriguez, E.~Osaba, J.~Poyatos,
  S.~Tabik, D.~Molina, and F.~Herrera, ``Lights and shadows in evolutionary
  deep learning: Taxonomy, critical methodological analysis, cases of study,
  learned lessons, recommendations and challenges,'' \emph{Information Fusion},
  vol.~67, pp. 161--194, 2021.

\bibitem[Langdon and Poli(2013)]{langdon2013foundations}
W.~B. Langdon and R.~Poli, \emph{Foundations of GeneticProgramming}.\hskip 1em
  plus 0.5em minus 0.4em\relax Springer Science \& Business Media, 2013.

\bibitem[Bi et~al.(2022{\natexlab{a}})Bi, Xue, and Zhang]{bi2021fewshot}
Y.~Bi, B.~Xue, and M.~Zhang, ``Dual-tree genetic programming for few-shot image
  classification,'' \emph{IEEE Trans. Evol. Comput.}, vol.~26, no.~3, pp.
  555--569, 2022.

\bibitem[Bi et~al.(2021{\natexlab{b}})Bi, Xue, and Zhang]{bi2021multi}
------, ``Multi-objective genetic programming for feature learning in face
  recognition,'' \emph{Appl. Soft Comput.}, vol. 103, p. 107152, 2021.

\bibitem[Bi et~al.(2021{\natexlab{c}})Bi, Xue, and Zhang]{bi2019tevc}
------, ``Genetic programming with image-related operators and a flexible
  program structure for feature learning in image classification,'' \emph{IEEE
  Trans. Evol. Comput.}, vol.~25, no.~1, pp. 87--101, 2021.

\bibitem[{Sun} et~al.(2019){Sun}, {Xue}, {Zhang}, and {Yen}]{sun2019evocnn}
Y.~{Sun}, B.~{Xue}, M.~{Zhang}, and G.~G. {Yen}, ``Evolving deep convolutional
  neural networks for image classification,'' \emph{IEEE Trans. Evol. Comput.},
  vol.~24, no.~2, pp. 1--14, 2019.

\bibitem[Sun et~al.(2019)Sun, Xue, Zhang, and Yen]{sun2019completely}
Y.~Sun, B.~Xue, M.~Zhang, and G.~G. Yen, ``Completely automated cnn
  architecture design based on blocks,'' \emph{IEEE Trans. Neural Netw. Learn.
  Syst.}, vol.~31, no.~4, pp. 1242--1254, 2019.

\bibitem[Lu et~al.(2020)Lu, Whalen, Dhebar, Deb, Goodman, Banzhaf, and
  Boddeti]{lu2020multiobjective}
Z.~Lu, I.~Whalen, Y.~Dhebar, K.~Deb, E.~D. Goodman, W.~Banzhaf, and V.~N.
  Boddeti, ``Multiobjective evolutionary design of deep convolutional neural
  networks for image classification,'' \emph{IEEE Trans. Evol. Comput.},
  vol.~25, no.~2, pp. 277--291, 2020.

\bibitem[Zhu and Jin(2022)]{zhu2021real}
H.~Zhu and Y.~Jin, ``Real-time federated evolutionary neural architecture
  search,'' \emph{IEEE Trans. Evol. Comput.}, vol.~26, no.~2, pp. 364--378,
  2022.

\bibitem[Zhang et~al.(2020)Zhang, Jin, Cheng, and Hao]{zhang2020efficient}
H.~Zhang, Y.~Jin, R.~Cheng, and K.~Hao, ``Efficient evolutionary search of
  attention convolutional networks via sampled training and node inheritance,''
  \emph{IEEE Trans. Evol. Comput.}, vol.~25, no.~2, pp. 371--385, 2020.

\bibitem[Rodriguez-Coayahuitl et~al.(2018)Rodriguez-Coayahuitl, Morales-Reyes,
  and Escalante]{rodriguez2018structurally}
L.~Rodriguez-Coayahuitl, A.~Morales-Reyes, and H.~J. Escalante, ``Structurally
  layered representation learning: Towards deep learning through genetic
  programming,'' in \emph{Proc. EuroGP}.\hskip 1em plus 0.5em minus 0.4em\relax
  Springer, 2018, pp. 271--288.

\bibitem[Shao et~al.(2014)Shao, Liu, and Li]{shao2014feature}
L.~Shao, L.~Liu, and X.~Li, ``Feature learning for image classification via
  multiobjective genetic programming,'' \emph{IEEE Trans. Neural Netw. Learn.
  Syst.}, vol.~25, no.~7, pp. 1359--1371, 2014.

\bibitem[Bi et~al.(2020)Bi, Xue, and Zhang]{bi2020effective}
Y.~Bi, B.~Xue, and M.~Zhang, ``An effective feature learning approach using
  genetic programming with image descriptors for image classification,''
  \emph{IEEE Comput. Intell. Mag.}, vol.~15, no.~2, pp. 65--77, 2020.

\bibitem[Bi et~al.(2021{\natexlab{d}})Bi, Xue, and Zhang]{bi2020genetic}
------, ``Genetic programming with a new representation to automatically learn
  features and evolve ensembles for image classification,'' \emph{IEEE Trans.
  Cybern.}, vol.~51, no.~4, pp. 1769--1783, 2021.

\bibitem[Bruintjes et~al.(2021)Bruintjes, Lengyel, Rios, Kayhan, and van
  Gemert]{bruintjes2021vipriors}
R.-J. Bruintjes, A.~Lengyel, M.~B. Rios, O.~S. Kayhan, and J.~van Gemert,
  ``Vipriors 1: Visual inductive priors for data-efficient deep learning
  challenges,'' \emph{arXiv preprint arXiv:2103.03768}, 2021.

\bibitem[Lengyel et~al.(2022)Lengyel, Bruintjes, Rios, Kayhan, Zambrano, Tomen,
  and van Gemert]{lengyel2022vipriors}
A.~Lengyel, R.-J. Bruintjes, M.~B. Rios, O.~S. Kayhan, D.~Zambrano, N.~Tomen,
  and J.~van Gemert, ``Vipriors 2: Visual inductive priors for data-efficient
  deep learning challenges,'' \emph{arXiv preprint arXiv:2201.08625}, 2022.

\bibitem[Zhang et~al.(2019)Zhang, Liu, Zhang, Zhang, and Zhu]{zhang2019deep}
L.~Zhang, J.~Liu, B.~Zhang, D.~Zhang, and C.~Zhu, ``Deep cascade model-based
  face recognition: When deep-layered learning meets small data,'' \emph{IEEE
  Trans. Image Process.}, vol.~29, pp. 1016--1029, 2019.

\bibitem[Arora et~al.(2019)Arora, Du, Li, Salakhutdinov, Wang, and
  Yu]{arora2019harnessing}
S.~Arora, S.~S. Du, Z.~Li, R.~Salakhutdinov, R.~Wang, and D.~Yu, ``Harnessing
  the power of infinitely wide deep nets on small-data tasks,'' in \emph{Proc.
  ICLR}, 2019.

\bibitem[Brigato and Iocchi(2021)]{brigato2021close}
L.~Brigato and L.~Iocchi, ``A close look at deep learning with small data,'' in
  \emph{Proc. IEEE ICPR}, 2021, pp. 2490--2497.

\bibitem[Bi et~al.(2022{\natexlab{b}})Bi, Xue, and Zhang]{bi2022using}
Y.~Bi, B.~Xue, and M.~Zhang, ``Using a small number of training instances in
  genetic programming for face image classification,'' \emph{Inf. Sci.}, 2022.

\bibitem[Barz and Denzler(2020)]{barz2020deep}
B.~Barz and J.~Denzler, ``Deep learning on small datasets without pre-training
  using cosine loss,'' in \emph{Proc. IEEE/CVF WCACV}, 2020, pp. 1371--1380.

\bibitem[Brigato et~al.(2021)Brigato, Barz, Iocchi, and
  Denzler]{brigato2021tune}
L.~Brigato, B.~Barz, L.~Iocchi, and J.~Denzler, ``Tune it or don't use it:
  Benchmarking data-efficient image classification,'' in \emph{Proc. IEEE
  ICCV}, 2021, pp. 1071--1080.

\bibitem[Sun et~al.(2020)Sun, Jin, Su, He, Xue, and Lu]{sun2020visual}
P.~Sun, X.~Jin, W.~Su, Y.~He, H.~Xue, and Q.~Lu, ``A visual inductive priors
  framework for data-efficient image classification,'' in \emph{Proc.
  ECCV}.\hskip 1em plus 0.5em minus 0.4em\relax Springer, 2020, pp. 511--520.

\bibitem[Zhao and Wen(2020)]{zhao2020distilling}
B.~Zhao and X.~Wen, ``Distilling visual priors from self-supervised learning,''
  in \emph{Proc. ECCV}.\hskip 1em plus 0.5em minus 0.4em\relax Springer, 2020,
  pp. 422--429.

\bibitem[Al-Sahaf et~al.(2016)Al-Sahaf, Zhang, and Johnston]{al2016binary}
H.~Al-Sahaf, M.~Zhang, and M.~Johnston, ``Binary image classification: A
  genetic programming approach to the problem of limited training instances,''
  \emph{Evol. Comput.}, vol.~24, no.~1, pp. 143--182, 2016.

\bibitem[Bi et~al.(2022{\natexlab{c}})Bi, Xue, and Zhang]{bi2020learning}
Y.~Bi, B.~Xue, and M.~Zhang, ``Learning and sharing: A multitask genetic
  programming approach to image feature learning,'' \emph{IEEE Trans. Evol.
  Comput.}, vol.~26, no.~2, pp. 218--232, 2022.

\bibitem[Montana(1995)]{montana1995strongly}
D.~J. Montana, ``Strongly typed genetic programming,'' \emph{Evol. Comput.},
  vol.~3, no.~2, pp. 199--230, 1995.

\bibitem[Dalal and Triggs(2005)]{dalal2005histograms}
N.~Dalal and B.~Triggs, ``Histograms of oriented gradients for human
  detection,'' in \emph{Proc. IEEE CVPR}, vol.~1, 2005, pp. 886--893.

\bibitem[Ojala et~al.(2002)Ojala, Pietikainen, and
  Maenpaa]{ojala2002multiresolution}
T.~Ojala, M.~Pietikainen, and T.~Maenpaa, ``Multiresolution gray-scale and
  rotation invariant texture classification with local binary patterns,''
  \emph{IEEE Trans. Pattern Anal. Mach. Intell.}, vol.~24, no.~7, pp. 971--987,
  2002.

\bibitem[Vedaldi and Fulkerson(2010)]{vedaldi2010vlfeat}
A.~Vedaldi and B.~Fulkerson, ``{VLFeat}: An open and portable library of
  computer vision algorithms,'' in \emph{Proc. 18th ACM Int. Conf. Multimedia},
  2010, pp. 1469--1472.

\bibitem[Krizhevsky et~al.(2009)Krizhevsky, Hinton,
  et~al.]{krizhevsky2009learning}
A.~Krizhevsky, G.~Hinton \emph{et~al.}, ``Learning multiple layers of features
  from tiny images,'' 2009.

\bibitem[Xiao et~al.(2017)Xiao, Rasul, and Vollgraf]{xiao2017fashion}
H.~Xiao, K.~Rasul, and R.~Vollgraf, ``Fashion-mnist: a novel image dataset for
  benchmarking machine learning algorithms,'' \emph{arXiv preprint
  arXiv:1708.07747}, 2017.

\bibitem[Netzer et~al.(2011)Netzer, Wang, Coates, Bissacco, Wu, and
  Ng]{netzer2011reading}
Y.~Netzer, T.~Wang, A.~Coates, A.~Bissacco, B.~Wu, and A.~Y. Ng, ``Reading
  digits in natural images with unsupervised feature learning,'' 2011.

\bibitem[Samaria and Harter(1994)]{samaria1994parameterisation}
F.~S. Samaria and A.~C. Harter, ``Parameterisation of a stochastic model for
  human face identification,'' in \emph{Proc. IEEE Workshop Appl. Comput.
  Vis.}, 1994, pp. 138--142.

\bibitem[Lee et~al.(2005)Lee, Ho, and Kriegman]{lee2005acquiring}
K.-C. Lee, J.~Ho, and D.~J. Kriegman, ``Acquiring linear subspaces for face
  recognition under variable lighting,'' \emph{IEEE Trans. Pattern Anal. Mach.
  Intell.}, no.~5, pp. 684--698, May 2005.

\bibitem[Liao et~al.(2021)Liao, Lei, Zhu, Zeng, Li, and Yuan]{liao2021deep}
T.~Liao, Z.~Lei, T.~Zhu, S.~Zeng, Y.~Li, and C.~Yuan, ``Deep metric learning
  for k nearest neighbour classification,'' \emph{IEEE Trans. Knowl. Data
  Eng.}, 2021.

\bibitem[He et~al.(2016)He, Zhang, Ren, and Sun]{he2016deep}
K.~He, X.~Zhang, S.~Ren, and J.~Sun, ``Deep residual learning for image
  recognition,'' in \emph{Proc. IEEE CVPR}, 2016, pp. 770--778.

\bibitem[Ulicny et~al.(2019)Ulicny, Krylov, and Dahyot]{ulicny2019harmonic222}
M.~Ulicny, V.~A. Krylov, and R.~Dahyot, ``Harmonic networks with limited
  training samples,'' in \emph{Proc. EUSIPCO}.\hskip 1em plus 0.5em minus
  0.4em\relax IEEE, 2019, pp. 1--5.

\bibitem[Simonyan and Zisserman(2014)]{simonyan2014very}
K.~Simonyan and A.~Zisserman, ``Very deep convolutional networks for
  large-scale image recognition,'' \emph{arXiv preprint arXiv:1409.1556}, 2014.

\bibitem[Zagoruyko and Komodakis(2016)]{zagoruyko2016wide}
S.~Zagoruyko and N.~Komodakis, ``Wide residual networks,'' \emph{arXiv preprint
  arXiv:1605.07146}, 2016.

\bibitem[Oyallon et~al.(2018)Oyallon, Zagoruyko, Huang, Komodakis,
  Lacoste-Julien, Blaschko, and Belilovsky]{oyallon2018scattering}
E.~Oyallon, S.~Zagoruyko, G.~Huang, N.~Komodakis, S.~Lacoste-Julien,
  M.~Blaschko, and E.~Belilovsky, ``Scattering networks for hybrid
  representation learning,'' \emph{IEEE Trans. Pattern Anal. Mach. Intell.},
  vol.~41, no.~9, pp. 2208--2221, 2018.

\bibitem[Masci et~al.(2011)Masci, Meier, Cire{\c{s}}an, and
  Schmidhuber]{masci2011stacked}
J.~Masci, U.~Meier, D.~Cire{\c{s}}an, and J.~Schmidhuber, ``Stacked
  convolutional auto-encoders for hierarchical feature extraction,'' in
  \emph{Proc. ICANN}.\hskip 1em plus 0.5em minus 0.4em\relax Springer, 2011,
  pp. 52--59.

\bibitem[Chen et~al.(2020)Chen, Sun, Zhang, and Peng]{chen2020evolving}
X.~Chen, Y.~Sun, M.~Zhang, and D.~Peng, ``Evolving deep convolutional
  variational autoencoders for image classification,'' \emph{IEEE Trans. Evol.
  Comput.}, vol.~25, no.~5, pp. 815--829, 2020.

\bibitem[Lezama et~al.(2018)Lezama, Qiu, Mus{\'e}, and Sapiro]{lezama2018ole}
J.~Lezama, Q.~Qiu, P.~Mus{\'e}, and G.~Sapiro, ``Ole: Orthogonal low-rank
  embedding-a plug and play geometric loss for deep learning,'' in \emph{Proc.
  IEEE CVPR}, 2018, pp. 8109--8118.

\bibitem[Bietti et~al.(2019)Bietti, Mialon, Chen, and Mairal]{bietti2019kernel}
A.~Bietti, G.~Mialon, D.~Chen, and J.~Mairal, ``A kernel perspective for
  regularizing deep neural networks,'' in \emph{Proc. ICML}.\hskip 1em plus
  0.5em minus 0.4em\relax PMLR, 2019, pp. 664--674.

\bibitem[Kayhan and Gemert(2020)]{kayhan2020translation}
O.~S. Kayhan and J.~C.~v. Gemert, ``On translation invariance in cnns:
  Convolutional layers can exploit absolute spatial location,'' in \emph{Proc.
  IEEE CVPR}, 2020, pp. 14\,274--14\,285.

\bibitem[Kobayashi(2021)]{kobayashi2021t}
T.~Kobayashi, ``T-vmf similarity for regularizing intra-class feature
  distribution,'' in \emph{Proc. IEEE CVPR}, 2021, pp. 6616--6625.

\bibitem[Turk and Pentland(1991)]{turk1991eigenfaces}
M.~Turk and A.~Pentland, ``Eigenfaces for recognition,'' \emph{J. Cogn.
  Neurosci.}, vol.~3, no.~1, pp. 71--86, 1991.

\bibitem[Belhumeur et~al.(1997)Belhumeur, Hespanha, and
  Kriegman]{belhumeur1997eigenfaces}
P.~N. Belhumeur, J.~P. Hespanha, and D.~J. Kriegman, ``Eigenfaces vs.
  fisherfaces: Recognition using class specific linear projection,'' \emph{IEEE
  Trans. Pattern Anal. Mach. Intell.}, vol.~19, no.~7, pp. 711--720, 1997.

\bibitem[He et~al.(2005{\natexlab{a}})He, Yan, Hu, Niyogi, and
  Zhang]{he2005face}
X.~He, S.~Yan, Y.~Hu, P.~Niyogi, and H.-J. Zhang, ``Face recognition using
  laplacianfaces,'' \emph{IEEE Trans. Pattern Anal. Mach. Intell.}, vol.~27,
  no.~3, pp. 328--340, 2005.

\bibitem[He et~al.(2005{\natexlab{b}})He, Cai, Yan, and
  Zhang]{he2005neighborhood}
X.~He, D.~Cai, S.~Yan, and H.-J. Zhang, ``Neighborhood preserving embedding,''
  in \emph{Proc. IEEE ICCV}, vol.~2.\hskip 1em plus 0.5em minus 0.4em\relax
  IEEE, 2005, pp. 1208--1213.

\bibitem[Yan et~al.(2006)Yan, Xu, Zhang, Zhang, Yang, and Lin]{yan2006graph}
S.~Yan, D.~Xu, B.~Zhang, H.-J. Zhang, Q.~Yang, and S.~Lin, ``Graph embedding
  and extensions: A general framework for dimensionality reduction,''
  \emph{IEEE Trans. Pattern Anal. Mach. Intell.}, vol.~29, no.~1, pp. 40--51,
  2006.

\bibitem[Zhang et~al.(2011)Zhang, Yang, and Feng]{zhang2011sparse}
L.~Zhang, M.~Yang, and X.~Feng, ``Sparse representation or collaborative
  representation: Which helps face recognition?'' in \emph{Proc. ICCV}, 2011,
  pp. 471--478.

\bibitem[Wright et~al.(2008)Wright, Yang, Ganesh, Sastry, and
  Ma]{wright2008robust}
J.~Wright, A.~Y. Yang, A.~Ganesh, S.~S. Sastry, and Y.~Ma, ``Robust face
  recognition via sparse representation,'' \emph{IEEE Trans. Pattern Anal.
  Mach. Intell.}, vol.~31, no.~2, pp. 210--227, 2008.

\bibitem[He et~al.(2010)He, Zheng, and Hu]{he2010maximum}
R.~He, W.-S. Zheng, and B.-G. Hu, ``Maximum correntropy criterion for robust
  face recognition,'' \emph{IEEE Trans. Pattern Anal. Mach. Intell.}, vol.~33,
  no.~8, pp. 1561--1576, 2010.

\bibitem[Yang et~al.(2011)Yang, Zhang, Yang, and Zhang]{yang2011robust}
M.~Yang, L.~Zhang, J.~Yang, and D.~Zhang, ``Robust sparse coding for face
  recognition,'' in \emph{Proc. IEEE CVPR}, 2011, pp. 625--632.

\bibitem[He et~al.(2013)He, Zheng, Tan, and Sun]{he2013half}
R.~He, W.-S. Zheng, T.~Tan, and Z.~Sun, ``Half-quadratic-based iterative
  minimization for robust sparse representation,'' \emph{IEEE Trans. Pattern
  Anal. Mach. Intell.}, vol.~36, no.~2, pp. 261--275, 2013.

\bibitem[Yang et~al.(2016)Yang, Luo, Qian, Tai, Zhang, and Xu]{yang2016nuclear}
J.~Yang, L.~Luo, J.~Qian, Y.~Tai, F.~Zhang, and Y.~Xu, ``Nuclear norm based
  matrix regression with applications to face recognition with occlusion and
  illumination changes,'' \emph{IEEE Trans. Pattern Anal. Mach. Intell.},
  vol.~39, no.~1, pp. 156--171, 2016.

\bibitem[Xie et~al.(2017)Xie, Yang, Qian, Tai, and Zhang]{xie2017robust}
J.~Xie, J.~Yang, J.~J. Qian, Y.~Tai, and H.~M. Zhang, ``Robust nuclear
  norm-based matrix regression with applications to robust face recognition,''
  \emph{IEEE Trans. Image Process.}, vol.~26, no.~5, pp. 2286--2295, 2017.

\bibitem[Yang et~al.(2014)Yang, Zhang, Feng, and Zhang]{yang2014sparse}
M.~Yang, L.~Zhang, X.~Feng, and D.~Zhang, ``Sparse representation based fisher
  discrimination dictionary learning for image classification,'' \emph{Int. J.
  Comput. Vis.}, vol. 109, no.~3, pp. 209--232, 2014.

\bibitem[Vu and Monga(2017)]{vu2017fast}
T.~H. Vu and V.~Monga, ``Fast low-rank shared dictionary learning for image
  classification,'' \emph{IEEE Trans. Image Process.}, vol.~26, no.~11, pp.
  5160--5175, 2017.

\bibitem[Bi et~al.(2019)Bi, Xue, and Zhang]{bi2019evolutionary}
Y.~Bi, B.~Xue, and M.~Zhang, ``An evolutionary deep learning approach using
  genetic programming with convolution operators for image classification,'' in
  \emph{Proc. IEEE CEC}, 2019, pp. 3197--3204.

\bibitem[Maaten and Hinton(2008)]{maaten2008visualizing}
L.~v.~d. Maaten and G.~Hinton, ``Visualizing data using t-sne,'' \emph{J. Mach.
  Learn. Res.}, vol.~9, pp. 2579--2605, 2008.

\end{thebibliography}

\end{document}